\documentclass[conference]{IEEEtran}
\IEEEoverridecommandlockouts
\usepackage{amsmath} % For additional math characters.
\usepackage{amssymb} % For additional math characters.

\usepackage{graphicx} % For advanced graphic files.
\usepackage{balance} % For balancing the columns.
\usepackage{amsfonts} % For additional math characters.
\usepackage{cite} % For additional citation formatting.
\usepackage[table]{xcolor} % For colored text/highlights.
\usepackage{bm} % For bold math argument.
\usepackage[hyphens]{url} % For breaking long URL.
\usepackage[hidelinks]{hyperref} % For internal/external links.
\usepackage[switch]{lineno} % For internal/external links.
\usepackage{tabularx} % For advanced tables.

\usepackage{textcomp}
\newcolumntype{C}{>{\centering\arraybackslash}X}
\newcolumntype{L}{>{\raggedright\arraybackslash}X}
\newcolumntype{R}{>{\raggedleft\arraybackslash}X}

\makeatletter
\let\MYcaption\@makecaption
\makeatother

\usepackage{subcaption}
\makeatletter
\let\@makecaption\MYcaption
\makeatother

\def\BibTeX{{\rm B\kern-.05em{\sc i\kern-.025em b}\kern-.08em
    T\kern-.1667em\lower.7ex\hbox{E}\kern-.125emX}}
\begin{document}

\title{
    Real-world~Instance-specific~Image~Goal~Navigation: Bridging Domain Gaps via Contrastive Learning
    % Contrastive Learning for Domain Gap Bridging in Image Goal Navigation for Service Robots
    % Real-world Instance-specific Image Goal Navigation for Service Robots: 
    % Bridging the Domain Gap with Contrastive Learning
    \thanks{
        This work was supported by JSPS KAKENHI Grants-in-Aid for Scientific Research (Grant Numbers JP23K16975, 22K12212) and JST Moonshot Research \& Development Program (Grant Number JPMJMS2011).
    }
}

\author{
    \IEEEauthorblockN{1\textsuperscript{st} Taichi Sakaguchi}
    \IEEEauthorblockA{\textit{Graduate School of Info. Sci. \& Eng.} \\
    \textit{Ritsumeikan University},
    Osaka, Japan \\
    sakaguchi.taichi@em.ci.ritsumei.ac.jp}
    \and
    \IEEEauthorblockN{2\textsuperscript{nd} Akira Taniguchi$^{*}$}
    \IEEEauthorblockA{\textit{College of Info. Sci. \& Eng.} \\
    \textit{Ritsumeikan University},
    Osaka, Japan \\
    a.taniguchi@em.ci.ritsumei.ac.jp}
    \and
    \IEEEauthorblockN{3\textsuperscript{rd} Yoshinobu Hagiwara$^{\dagger}$}
    \IEEEauthorblockA{\textit{Faculty of Sci. and Eng.} \\
    \textit{Soka University},
    Tokyo, Japan \\
    % \textit{Research Organization of Sci. \& Tech.}\\
    % \textit{Ritsumeikan University}, Shiga, Japan \\
    hagiwara@soka-u.jp}
    % yhagiwara@em.ci.ritsumei.ac.jp}
    \and
    \IEEEauthorblockN{4\textsuperscript{th} Lotfi El Hafi}
    \IEEEauthorblockA{\textit{Research Organization of Sci. \& Tech.} \\
    \textit{Ritsumeikan University},
    Shiga, Japan \\
    lotfi.elhafi@em.ci.ritsumei.ac.jp}
    \and
    \IEEEauthorblockN{5\textsuperscript{th} Shoichi Hasegawa}
    \IEEEauthorblockA{\textit{Graduate School of Info. Sci. \& Eng.} \\
    \textit{Ritsumeikan University},
    Osaka, Japan \\
    hasegawa.shoichi@em.ci.ritsumei.ac.jp}
    \and
    \IEEEauthorblockN{6\textsuperscript{th} Tadahiro Taniguchi$^{\dagger}$}
    \IEEEauthorblockA{\textit{Graduate School of Informatics} \\
    \textit{Kyoto University},
    Kyoto, Japan \\
    % \textit{Research Organization of Sci. \& Tech.}\\
    % \textit{Ritsumeikan University}, Shiga, Japan \\
    taniguchi@i.kyoto-u.ac.jp} % taniguchi@em.ci.ritsumei.ac.jp
    % \and
    % \IEEEauthorblockN{Tadahiro Taniguchi}
    % \IEEEauthorblockA{
    % \textit{Graduate School of Informatics}, 
    % \textit{Kyoto University}, Kyoto, Japan}
    % \textit{Research Organization of Sci. \& Tech.},
    % \textit{Ritsumeikan University}, Shiga, Japan \\
    % taniguchi@i.kyoto-u.ac.jp, taniguchi@em.ci.ritsumei.ac.jp
    \thanks{
        $^{*}$Corresponding author.
    }
    \thanks{
        $^{\dagger}$ They are also affiliated with the Research Organization of Sci. \& Tech., Ritsumeikan University, Shiga, Japan.
        %Thay are also with Research Organization of Sci. \& Tech., Ritsumeikan University, Shiga, Japan.
    }
}

%%%%%%%%%%%%%%%%%%%%%%%%%%%%%%%%%%%%%%%%%%%%%%%%%%%%%%%%%%%%%%%%%%%%%%%%%%%%%%%%

% \title{\LARGE\bf
%     Real-world Instance-specific Image Goal Navigation for Service Robots:\\
%     Bridging the Domain Gap with Contrastive Learning
% }

% \author{
%     Taichi Sakaguchi$^{1}$,
%     Akira Taniguchi$^{1, *}$,
%     Yoshinobu Hagiwara$^{1}$,
%     Lotfi El Hafi$^{1}$,\\
%     Shoichi Hasegawa$^{1}$, and
%     Tadahiro Taniguchi$^{1}$
%     \thanks{
%         This work was supported by JSPS KAKENHI Grants-in-Aid for Scientific Research (Grant Numbers JP23K16975, 22K12212) and JST Moonshot Research \& Development Program (Grant Number JPMJMS2011).
%     }
%     \thanks{
%         $^{1}$Taichi Sakaguchi, Akira Taniguchi, Yoshinobu Hagiwara, Lotfi El Hafi, Shoichi Hasegawa, and Tadahiro Taniguchi are with Ritsumeikan University;
%         1-1-1 Noji-Higashi, Kusatsu, Shiga 525-8577, Japan.
%         {\tt\small\{sakaguchi.taichi, a.taniguchi, yhagiwara, lotfi.elhafi, hasegawa.shoichi, taniguchi\} @em.ci.ritsumei.ac.jp}
%     }
%     \thanks{
%         $^{*}$Corresponding author.
%     }
% }

%%%%%%%%%%%%%%%%%%%%%%%%%%%%%%%%%%%%%%%%%%%%%%%%%%%%%%%%%%%%%%%%%%%%%%%%%%%%%%%%
% Abstract
%%%%%%%%%%%%%%%%%%%%%%%%%%%%%%%%%%%%%%%%%%%%%%%%%%%%%%%%%%%%%%%%%%%%%%%%%%%%%%%%

\maketitle

% Default template styles for final submission:
% \thispagestyle{empty}
% \pagestyle{empty}

% Temporary styles for displaying line numbers:
\pagestyle{plain}
% \linenumbers %arxivで使えないので停止

%%%%%%%%%%%%%%%%%%%%%%%%%%%%%%%%%%%%%%%%%%%%%%%%%%%%%%%%%%%%%%%%%%%%%%%%%%%%%%%%

% ロボットシステムがユーザーの望む物体を見つけるのを支援するためには、クエリ画像から実世界の環境で同一の物体を特定する「インスタンス特定型画像ナビゲーション（InstanceImageNav）」の改善が重要です。しかし、移動中のロボットが観測する低品質な画像（動きによるブレや低解像度が特徴）と、ユーザーが提供する高品質なクエリ画像との間に存在する「ドメインギャップ」が課題となります。このドメインギャップはタスク成功率を大幅に低下させる可能性があるにもかかわらず、これまでの研究では十分に注目されていませんでした。
% この問題を解決するために、私たちは「CrossIA（few-shot cross-quality instance-aware adaptation）」と呼ばれる新しい手法を提案します。CrossIAは、コントラスト学習とインスタンス分類器を用いて、大量の低品質画像と少量の高品質画像との間で特徴を整合させることで、ドメインギャップを効果的に縮小します。このアプローチにより、クロス品質の画像間での潜在表現をインスタンス単位で近づけることが可能になります。
% さらに、システムは観測された画像の品質を向上させるために、事前に学習されたデブレモデルを使用して、物体画像コレクションを統合します。私たちの手法では、ImageNetで事前学習されたSimSiamモデルをCrossIAを用いてファインチューニングします。
% 提案手法の有効性は、20種類のインスタンスを対象としたInstanceImageNavタスクで評価しました。このタスクでは、ロボットが実世界の環境で高品質なクエリ画像と同じインスタンスを特定します。実験の結果、提案手法は、従来のSuperGlueに基づくアプローチと比較して、タスク成功率を最大で3倍に向上させることが示されました。
% これらの結果は、コントラスト学習と画像強化技術を活用してドメインギャップを橋渡しし、ロボティックアプリケーションにおける物体ローカリゼーションの向上に寄与する可能性を示しています。
% プロジェクトのウェブサイトは、こちらです。
\begin{abstract}
    Improving instance-specific image goal navigation (InstanceImageNav), which involves locating an object in the real world that is identical to a query image, is essential for enabling robots to help users find desired objects.
    The challenge lies in the domain gap between the low-quality images observed by the moving robot, characterized by motion blur and low resolution, and the high-quality query images provided by the user.
    These domain gaps can significantly reduce the task success rate, yet previous work has not adequately addressed them.
    To tackle this issue, we propose a novel method: few-shot cross-quality instance-aware adaptation (CrossIA). This approach employs contrastive learning with an instance classifier to align features between a large set of low-quality images and a small set of high-quality images.
    We fine-tuned the SimSiam model, pre-trained on ImageNet, using CrossIA with instance labels based on a 3D semantic map. 
    Additionally, our system integrates object image collection with a pre-trained deblurring model to enhance the quality of the observed images.
    Evaluated on an InstanceImageNav task with 20 different instance types, our method improved the task success rate by up to three-fold compared to a baseline based on SuperGlue. 
    These findings highlight the potential of contrastive learning and image enhancement techniques in improving object localization in robotic applications. 
    The project website is \href{https://emergentsystemlabstudent.github.io/DomainBridgingNav/}{\textcolor{blue}{https://emergentsystemlabstudent.github.io/DomainBridgingNav/}}.
\end{abstract}


\begin{figure}[t]
    \centering
    \includegraphics[width=\linewidth]{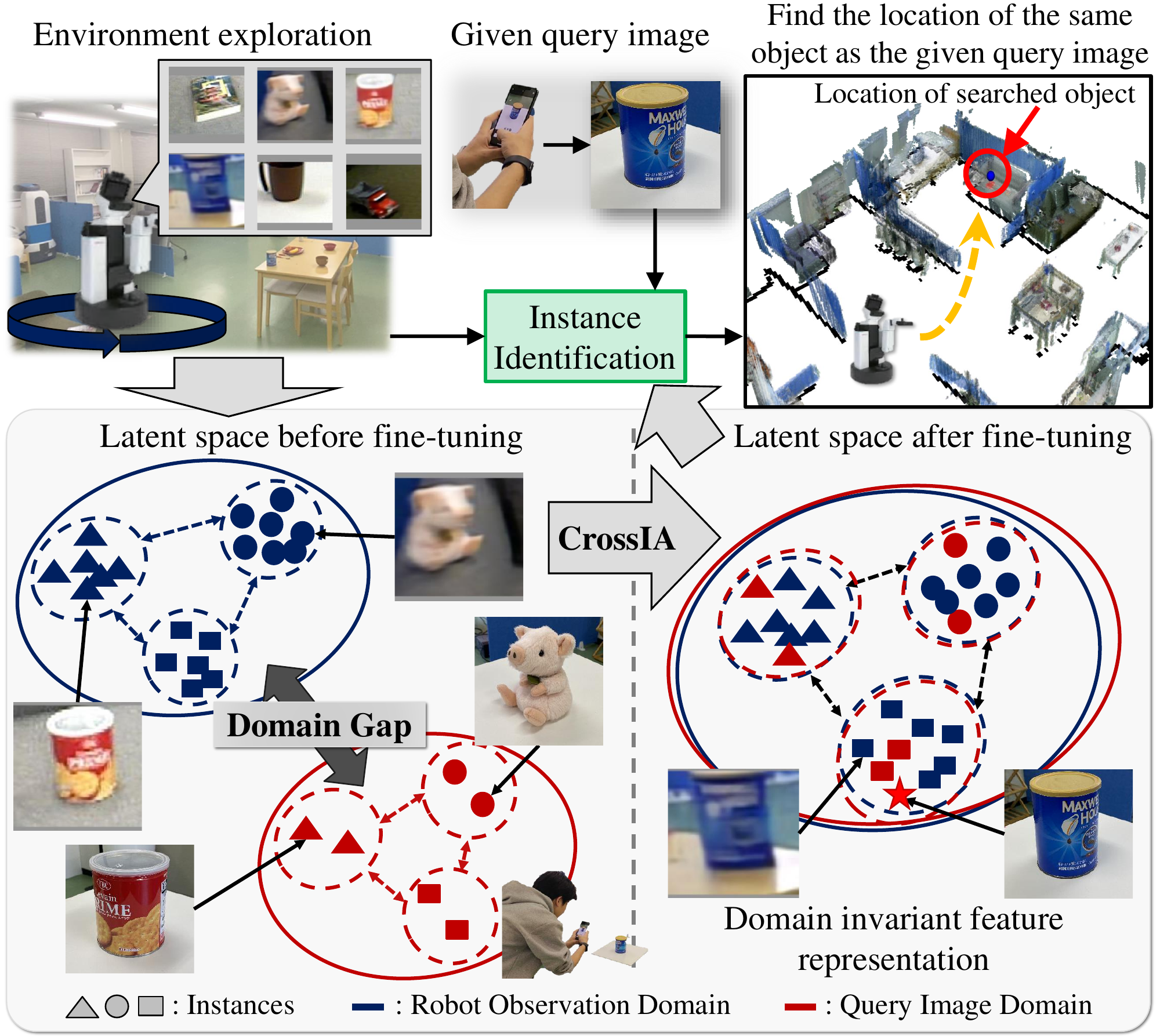}
    \caption{
        % 本研究のフォーカスタスク。
        % (下）ユーザの携帯電話から提供されたクエリ画像に写っている物体の位置を特定するロボット。
        % (左上）ユーザの携帯電話で撮影された画像と、実際のロボットが観察した物体画像とでは、画質が大きく異なるというドメインのギャップ。
        % (右上）画質の異なる同一インスタンスの画像を潜在空間で整列させる対比学習。
        Focused task in this study.
        (Top) The robot identifies the position of an object shown in a query image provided by a user's mobile phone.
        (Bottom left) Domain gap that the image quality significantly differs between the image taken by the user's mobile phone and the object image observed by the real robot.
        (Bottom right) Contrastive learning to align images of the same instance with different image quality in latent space.
    }
    \label{fig:overview_instanceimagenav}
\end{figure}

% \newpage
%%%%%%%%%%%%%%%%%%%%%%%%%%%%%%%%%%%%%%%%%%%%%%%%%%%%%%%%%%%%%%%%%%%%%%%%%%%%%%%%
% Introduction
%%%%%%%%%%%%%%%%%%%%%%%%%%%%%%%%%%%%%%%%%%%%%%%%%%%%%%%%%%%%%%%%%%%%%%%%%%%%%%%%

\section{Introduction}
\label{sec:introduction}

% 複雑な実世界の家庭環境において、ロボットが特定のインスタンスを効率的に 探索するためには、〚インスタンスに特化した画像ゴールナビゲーション (InstanceImageNav)〛を解くことが重要である。
% このタスクでは、図～ref{fig:overview_instanceimagenav}（下）に示すように、ロボットはユーザから提供されたクエリ画像に示されたオブジェクトを見つける。
% 同じオブジェクトクラスの複数のインスタンスが環境内に存在し、それぞれが異なる外観を持つ場合、この課題はさらに複雑になる。
% このようなケースでタスクを成功させるためには、ロボットは同じクラスのオブジェクトの異なるインスタンスを識別する必要がある。
% ターゲットとなるインスタンスと他のインスタンスが視覚的に区別できれば、ロボットは提供されたクエリ画像と同じオブジェクトを見つけることができる。
The importance of solving \textbf{instance-specific image goal navigation (InstanceImageNav)}~\cite{instanceimagenav} for robots to effectively search for specific instances in complex real-world domestic environments.
In this task, the robot locates an object in a query image provided by the user, as shown in Fig.~\ref{fig:overview_instanceimagenav} (top).
This task becomes even more difficult when multiple instances of the same object class, each with a distinct appearance, are present in the environment.
To successfully execute tasks in such scenarios, robots must be able to distinguish between different instances of the same class of objects.
If the target instance can be visually differentiated from others, the robot can successfully locate an object identical to the provided query image.

% In this task, the robot locates the object shown in the query image provided by the user, as shown in Fig.~\ref{fig:overview_instanceimagenav} (top).
% This challenge is further compounded when multiple instances of the same object class, each with a different appearance, are present within the environment.
% To successfully execute tasks in such cases, robots need to identify different instances of the same class of objects.
% If the target instance and other instances can be visually distinguished, the robot can locate an object identical to the provided query image.

\begin{figure}[t]
    \centering
    \includegraphics[width=0.9\linewidth]{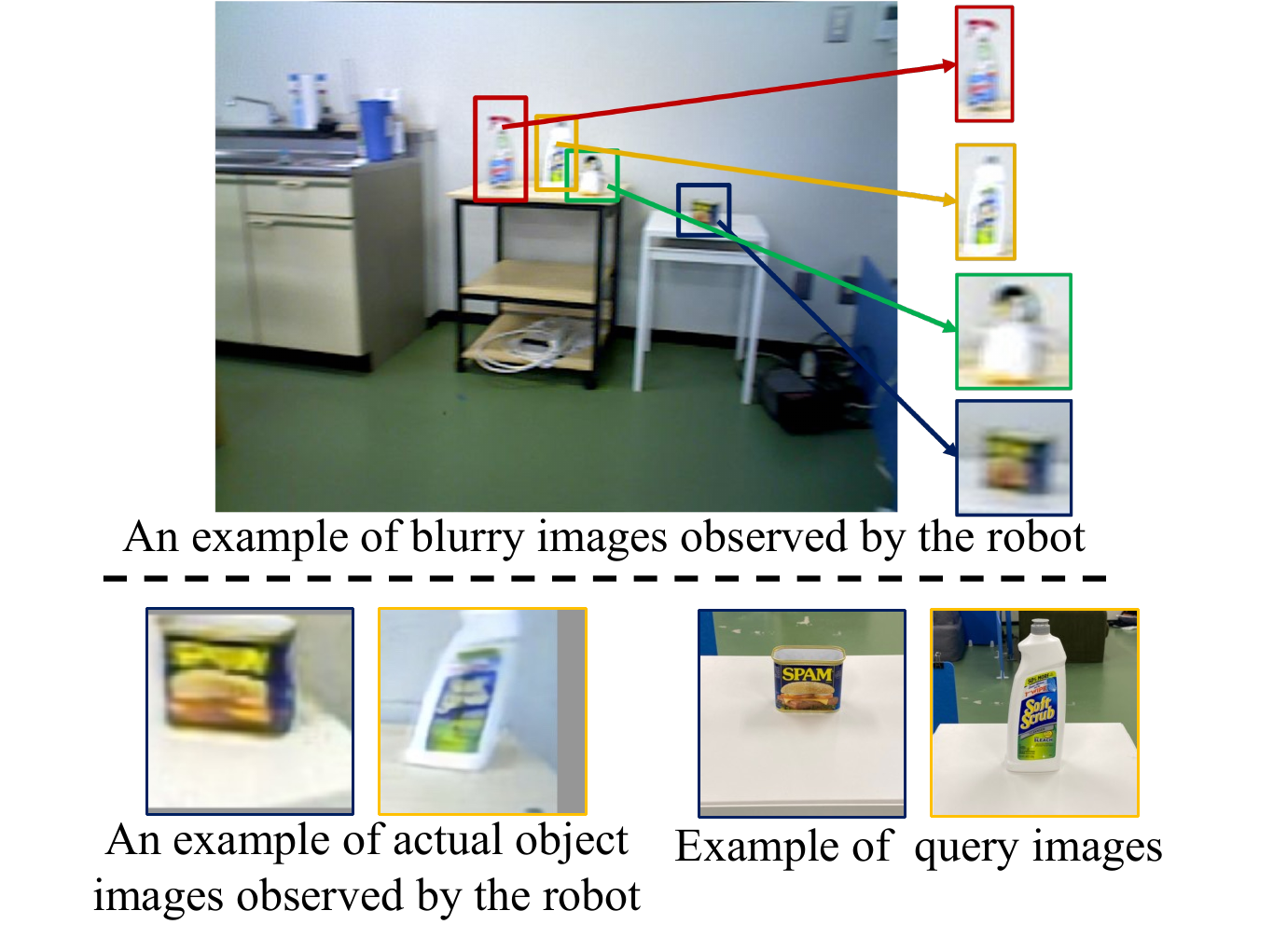}
    \caption{
        (Top) Object images cropped from the robot's observation image;
        (Bottom left) Examples of low-quality images;
        (Bottom right) Examples of high-quality images as the query.
    }
    \label{fig:domain_gap_example}
\end{figure}

% 実世界における課題は、ロボットが撮影する低画質画像と、ユーザーが携帯電話などのデバイスを通じて提供する一般的な高画質画像との間に内在するドメインギャップにある。
% このような画質の不一致は、InstanceImageNavの成功率を著しく低下させる。
% ロボットが環境を探索するとき、観察される画像は、動きによるモーションブラーや低解像度のために低画質であることが多い。
% 低画質の原因は、画角の中で物体が小さく見えることである（図～refig:domain_gap_example}（上）参照）。
% この問題は、画角と画質の両方が不十分な、家庭で使用される民生用ロボットによって収集されたデータで特に顕著である。
% したがって、図～ref{fig:overview_instanceimagenav}（上）に示すように、ロボットの観察画像とユーザが提供するクエリ画像の間のドメインギャップを埋めることが重要である。
%加えて、ロボットは大量に画像を得られるが、人の画像は数枚程度である。
The challenge in real-world environments is the \textbf{domain gap} between the low-quality images captured by the robot and the high-quality images provided by users (see Fig.~\ref{fig:overview_instanceimagenav} (bottom left)).
This discrepancy in image quality significantly reduces the success rate of InstanceImageNav.
When robots explore an environment, the images they capture are often low-quality due to motion blur, low resolution from movement, and the limitations of their onboard sensors.
 Objects may appear small in the robot's field of view (see Fig.~\ref{fig:domain_gap_example} (top)), further reducing image quality.
This issue is particularly pronounced in data collected by consumer robots used in a domestic setting, where the field angle of view and image quality may be insufficient.
Hence, it is important to bridge the domain gap between the robot's observed images and the query images provided by the user (see Fig.~\ref{fig:overview_instanceimagenav} (bottom right)).
 Additionally, while the robot can collect many images, the user can only provide a limited number of environment-specific images without significant effort.

% この課題を解決するために、我々は2つの重要なメカニズムを統合したシステムを提案します。1つは、「大量の低画質画像と少数の高画質画像の間で不変な特徴表現を学習する 」ことであり、もう1つは、ロボットが観察する画像の画質を向上させることです。
To address this challenge, we propose a system that integrates two key mechanisms: one for \textbf{learning invariant feature representations between massive low-quality and a few high-quality images} and another for enhancing the quality of images observed by robots using a pre-trained deblurring model~\cite{MSSNet}.
This system is based on an object image database construction system~\cite{goat}.
% 本研究では、インスタンス分類器による対比学習を通して、「数ショット交差品質インスタンス認識適応(CrossIA) 」という手法を提案する。
For the first mechanism, we propose a \textbf{few-shot cross-quality instance-aware adaptation (CrossIA)} through contrastive learning with an instance classifier.
% contrastive learning
% 「対比学習 」は、異なるドメイン間の同一インスタンスの画像間で不変な特徴表現を獲得することを可能にする。
Contrastive learning enables the acquisition of invariant feature representations between images of the same instance in different domains~\cite{contrastive_uda, contrastive_learning_domain_adaptation}.
% This system is based on an object image database construction system~\cite{goat}.
% 環境探索によって構築された3次元セマンティックマップは、自動的にデータベースにインスタンスラベルを提供する
The 3D semantic map constructed through environment exploration automatically provides instance labels to the database.
% To mitigate motion blur, we integrate a pre-trained deblurring model~\cite{MSSNet} into an object image database construction system similar to the system of Chang~\textit{et~al.}~\cite{goat}.
% 本研究では、ロボットが実環境を探索しながら収集した物体画像から、クエリ画像と同一の物体を特定する提案システムの有効性を評価する。
% これまでのInstanceImageNavの研究では、1つの物体に対して1つのクエリ画像しか撮影されていなかった~\cite{modular_instancenav, goat}。
% しかし、画像はユーザによって様々な角度から撮影される可能性がある。
% そこで、提案システムを様々な角度から評価するために、各インスタンスに対して8枚の画像を収集した。
% 本研究では、配置されるオブジェクトはテーブル上に配置可能な20種類のインスタンスである。
% 我々の主な貢献は2つある。
% \begin{列挙｝
%     \item我々は、インスタンス分類器による対比学習が不可欠であることを示し、少数ショット条件下で、異質な画像間の不変な特徴表現を学習し、既存のデブラーリング技術だけでは解決できない領域ギャップを効果的に埋める。
%     \項目我々は、3Dセマンティックマッピングを含むデータ収集モジュール、事前に学習されたデブラーリングモデル、およびSimSiamモデルの微調整を統合した提案システムが、InstanceImageNavタスクにおいて高い精度を達成できることを示す。
% \終わり
We evaluated the performance in identifying the object identical to the query image from the object images collected by the robot while exploring the real-world environment.
% In previous work on InstanceImageNav, only one query image was taken for each object~\cite{modular_instancenav, goat}.
% However, images can be captured from various angles by the user.
% Therefore, we collected eight images for each instance to evaluate the proposed system from various angles of the query object.
% In this study, 
The located objects are 20 different instances that can be placed on a table.
Our main contributions are two-fold.
\begin{enumerate}
    \item  We show that combining contrastive learning with few-shot learning and an instance classifier is essential for learning invariant feature representations between cross-quality images under few-shot conditions, effectively bridging the domain gap that cannot be resolved by deblurring alone.
    % We show that combining contrastive learning with few-shot learning and an instance classifier is essential, to learn invariant feature representations between cross-quality images under few-shot conditions, effectively bridging the domain gap that cannot be resolved solely by existing deblurring.
    % We show that contrastive learning with an instance classifier is essential, to learn invariant feature representations between cross-quality images under few-shot conditions, effectively bridging the domain gap that cannot be resolved solely by existing deblurring
    \item We experimentally demonstrate that the proposed method improves task success rates by up to three-times better than existing methods, SuperGlue~\cite{superglue}, highlighting its potential to identify small everyday objects in real-world InstanceImageNav.
    % We experimentally demonstrate that the proposed method improves task success rates by up to three-fold compared to existing methods, SuperGlue~\cite{superglue}, highlighting its potential to small everyday objects in real-world InstanceImageNav.
    % We show that the proposed system, integrating a data collection module involving 3D semantic mapping, a pre-trained deblurring model, and fine-tuning the SimSiam model, can achieve high accuracy in InstanceImageNav tasks.
\end{enumerate}

% 本稿の残りの部分は以下のように構成されている。
% セクション~ref{sec:related_works}で関連作品の概要を述べる。
% セクション~ref{sec:proposed_system}で提案手法の統合を詳述する。
% sec:experiments}節で、実験プロトコルと測定基準について述べる。
% さらに、セクション~ref{sec:results}で結果を提示して議論する。
% 最後に、セクション~ref{sec:conclusion}で今後の研究の方向性について結論を述べる。
The remainder of this paper is organized as follows.
Section~\ref{sec:problem} presents the problem statement, detailing the specific challenges addressed by our study. 
Section~\ref{sec:related_works} provides an overview of the related work.
Section~\ref{sec:proposed_system} describes the proposed method.
Section~\ref{sec:experiments} describes our experimental protocols and metrics.
Section~\ref{sec:results} presents and discusses our results.
Finally, Section~\ref{sec:conclusion} concludes with directions for future work.

%%%%%%%%%%%%%%%%%%%%%%%%%%%%%%%%%%%%%%%%%%%%%%%%%%%%%%%%%%%%%%%%%%%%%%%%%%%%%%%%
% Problem Statement
%%%%%%%%%%%%%%%%%%%%%%%%%%%%%%%%%%%%%%%%%%%%%%%%%%%%%%%%%%%%%%%%%%%%%%%%%%%%%%%%

\section{Problem Statement}
\label{sec:problem}

% 本研究は、現実世界のInstanceImageNavを対照学習で発展させるために不可欠な重要課題に取り組んだ。
This study addressed the essential challenges for the advancement of InstanceImageNav in the real world.

    % 画像品質の格差によるドメインギャップ：
    % ロボットが撮影した低画質の画像と、ユーザーが提供した高画質のクエリ画像との間のドメインギャップは、第一の障害である。このギャップは、モーションブラー、低解像度、およびロボットが動いているときに画質を劣化させるその他の要因によって生じる。その結果、ロボットがこれらの画像を高品質なクエリ画像と照合する能力が著しく損なわれる。
    \textbf{Domain gap by image quality disparity}: 
    The domain gap between the low-quality images captured by robots and the high-quality query images provided by users is a major challenge. 
    This gap stems from motion blur, low resolution, and other factors that diminish image quality when robots are in motion. 
    Consequently, the robot's capability to match these images with high-quality query images is greatly hindered.
    % The domain gap between the low-quality images captured by robots and the high-quality query images provided by users is a primary obstacle. This gap arises due to motion blur, low resolution, and other factors that degrade image quality when robots are in motion. 
    % As a result, the robot's ability to match these images with high-quality query images is significantly impaired.
    % 
    % ロボットは、動きやセンサーの制約から、ブレや解像度の低い画像を撮影することが多い。一方、ユーザーからのクエリ画像は通常、高画質で鮮明、かつ明るいため、2つの画像セットの間に大きなミスマッチが生じる。
    % Robots often capture images with motion blur and low resolution because of movement and sensor limitations. In contrast, query images from users are usually high-quality, clear, and well-lit, leading to a substantial mismatch between the two sets of images.

    % 既存手法のロバストネスの限界： InstanceImageNavの現在のアプローチは、多くの場合、大きくて識別しやすい物体に焦点を当てたり、動きのブレや解像度の低さなど、実環境の複雑さを十分にとらえられないシミュレーションに依存しています。このような手法は、通常、高画質画像の大規模なデータセットが利用可能であることを前提としていますが、実際のアプリケーションでは必ずしも実現可能ではありません。
    \textbf{Limited robustness of previous work}: 
    % 現在のインスタンスイメージナビアプローチのほとんどはシミュレーション環境で開発、テストされており、モーションブラーのような現実世界の課題を考慮できていません。
    % さらに、InstanceImageNav~~\cite{instanceimagenav, modular_instancenav, goat}の先行研究では、Ⓐテキストtt{椅子}、Ⓐテキストtt{机}、Ⓑテキストtt{テレビ}のような、大きくて区別しやすい物体を対象とすることが多い。
    % しかし、ロボットはより小さい物体や視覚的に類似した物体を見つける必要がある場合があり、そのような物体は実用的なシナリオではより一般的である可能性があります。
    % このような物体の画像を収集するとき、解像度が低くなりがちで、画質がユーザが 提供したクエリ画像と異なることがある（図～ref{fig:domain_gap_example}（下）参照）。
    Most current InstanceImageNav approaches have been developed and tested in the Habitat Simulator~\cite{habitat_sim}, which does not account for real-world challenges such as motion blur~\cite{instanceimagenav, modular_instancenav, sakaguchi_iros2024}.     
    Furthermore, previous work on InstanceImageNav~\cite{instanceimagenav, modular_instancenav, goat} often targets large and easily distinguishable objects, such as \texttt{chair}, \texttt{desk}, and \texttt{television}.
    However, robots might need to locate smaller or visually similar objects that are more common in practical scenarios, such as \texttt{cups}, \texttt{bottles}, and \texttt{books}.
    When collecting images of such objects, the resolution tends to be low and image quality may differ from the query image provided by the user (see Fig.~\ref{fig:domain_gap_example} (bottom)).

%%%%%%%%%%%%%%%%%%%%%%%%%%%%%%%%%%%%%%%%%%%%%%%%%%%%%%%%%%%%%%%%%%%%%%%%%%%%%%%%
% Related Work
%%%%%%%%%%%%%%%%%%%%%%%%%%%%%%%%%%%%%%%%%%%%%%%%%%%%%%%%%%%%%%%%%%%%%%%%%%%%%%%%

\section{Related Work}
\label{sec:related_works}

%%%%%%%%%%%%%%%%%%%%%%%%%%%%%%%%%%%%%%%%%%%%%%%%%%%%%%%%%%%%%%%%%%%%%%%%%%%%%%%%

\subsection{Object Searching}

% オブジェクト検索に関する研究は、検索対象の表現によって3つのタイプに分類される。
% 第一に、オブジェクトゴールナビゲーション(ObjectNav)は、``book''のような与えられたクラス名に属するオブジェクトを探す。
% 次に、視覚言語ナビゲーション(VLN: vision-and-language navigation)は、「リビングルームに行って、四角いテーブルの上にある黄色いコップを取りなさい」というような言語指示によって表現される物体の位置を特定する。
% 従って、VLN法は特定のインスタンスを検索することも可能である~\cite{multi_rankit}。
% しかし、これらの言語指示は、ユーザが対象オブジェクトが環境内のどこに位置するかを理解する必要がある。
% ユーザが目的のオブジェクトの位置を知らない場合、ユーザが過去に撮影したクエリ画像と同一のオブジェクトを発見する能力が重要になる。
% したがって、特定のインスタンスを検索する機能を開発するためには、InstanceImageNavへの対応が不可欠である。
The search for objects is classified into three types based on the representation of the search target.
First, object goal navigation (ObjectNav) involves locating any object belonging to a given class,  such as \texttt{book}~\cite{survey_objectnav}.
Second, vision-and-language navigation (VLN) involves locating an object represented by language instructions such as ``Go to the living room and pick up the yellow cup on the square table''~\cite{survey_vln, multi_rankit}. 
This method also enables the search for specific instances~\cite{multi_rankit}.
However, these language instructions require the user to know where the target object is located within the environment.
When the user is unaware of the object's location, the ability to locate an object identical to a query image captured by the user becomes crucial. 
Therefore, addressing InstanceImageNav is essential for developing the capability to search for specific instances

% 物体探索の方法は大きく2種類に分類できる。
% 一つは、深層強化学習や模倣学習に基づいて、ロボットの観察画像から直接行動を生成するニューラルネットワークを学習するend-to-end方式である~\cite{instanceimagenav, survey_objectnav, ovrl}。
% もう一つは、物体認識やSLAMなどのコンポーネントを統合したモジュール方式である~\cite{modular_instancenav, goat, survey_objectnav}。
% end-to-end法は、サンプルの非効率性~~instanceimagenav}とシムからリアルへの転送の悪さ~~realworld_objectnav}に悩まされる。
% しかし、モジュール方式では、実世界でもシミュレーションと同じ成功率でタスクを実行することができる〜cite{realworld_objectnav}。
% そこで、モジュール方式のナビゲーションシステムも提案する。
Methods for object searching can be broadly categorized into two types.
One type is the end-to-end method, which involves learning neural networks that directly generate actions from the robot's observed images using deep reinforcement learning or imitation learning~\cite{instanceimagenav, survey_objectnav, ovrl}.
The other type is the modular method, which integrates components such as pre-trained object recognition and simultaneous localization and mapping (SLAM)~\cite{modular_instancenav, goat, survey_objectnav, sakaguchi_iros2024}.
The end-to-end method suffers from sample inefficiency~\cite{instanceimagenav} and poor sim-to-real transfer~\cite{realworld_objectnav}.
Therefore, we also propose a modular navigation system.

% However, the modular method enables tasks to be performed with the same success rate in the real world as in simulation~\cite{realworld_objectnav}.

% InstanceImageNavを解決するために、先行研究は、事前に訓練された画像認識モデルやSLAM（Simultaneous Localization and Mapping）など、複数の技法~\cite{modular_instancenav, goat}を組み合わせたナビゲーションシステムを提案している。
% 例えば、Chang~textit{et~al.}は、環境を探索し、各インスタンスの3次元意味マップと画像を保存するデータベースを構築することで、InstanceImageNavタスクを解決するシステムを提案した~\cite{goat}。
% 彼らは、局所特徴マッチングの手法であるSuperGlue~\cite{superglue}を用いてInstanceImageNavを実行し、与えられたクエリ画像と同じ物体をデータベースから探索し、3次元意味地図上の位置を特定する。
To solve InstanceImageNav, previous work proposed navigation systems that combine multiple techniques~\cite{modular_instancenav, goat, sakaguchi_iros2024}.
% , such as a pre-trained image recognition model and SLAM.
For instance, Chang~\textit{et~al.} proposed a system exploring the environment and constructing a database that saves 3D semantic maps and images of each instance~\cite{goat}.
They execute InstanceImageNav using the local feature matching, SuperGlue~\cite{superglue}, to search for the same object as the given query image from the database and identify its position on the 3D semantic map.

%%%%%%%%%%%%%%%%%%%%%%%%%%%%%%%%%%%%%%%%%%%%%%%%%%%%%%%%%%%%%%%%%%%%%%%%%%%%%%%%

\subsection{Contrastive Learning}

% 対比学習とは、自己教師学習の一種で、同じインスタンスの異なる画像を対比するタスクから学習することである~\cite{simclr, moco, simsiam, byol}。
% 対比学習で事前に訓練されたモデルは、同じクラスに属する異なるインスタンスを識別する識別的な特徴表現を獲得することが知られている~\cite{reverse_engineering}。
% 例えば、Ido~textit{et~al.}は、対比学習とk-nearest neighbor法で学習された画像エンコーダから抽出された特徴量を用いて、インスタンスレベルとクラスレベルの画像分類タスクで高い精度を示した~\cite{reverse_engineering}。
% したがって、対比学習によって事前に訓練されたモデルは、InstanceImageNavのような、クエリ画像と同一のオブジェクトを見つけることを目的とするタスクに適していると考えられる。
Contrastive learning is a form of self-supervised learning that involves learning from tasks that contrast different images of the same instance~\cite{simclr, moco, simsiam, byol}. 
Models pre-trained with contrastive learning are known for acquiring discriminative feature representations that effectively differentiate instances within the same class~\cite{reverse_engineering}. 
For instance, Ido~\textit{et~al.} demonstrated high precision in image classification tasks at both the instance and class levels using features extracted from an image encoder trained with contrastive learning, combined with the k-nearest neighbor method~\cite{reverse_engineering}. 
Consequently, models trained via contrastive learning are deemed well-suited for the InstanceImageNav task.
% s such as InstanceImageNav, where the objective is to locate the object identical to the query image.

% SimView~\cite{sakaguchi_iros2024}
SimView improved performance on InstanceImageNav by fine-tuning a pre-trained SimSiam~\cite{simsiam} with contrastive learning using instance labels derived from a 3D semantic map~\cite{sakaguchi_iros2024}.
This approach forms the basis for this study.
However, it has not yet been tested for domain gaps on real robots.

%%%%%%%%%%%%%%%%%%%%%%%%%%%%%%%%%%%%%%%%%%%%%%%%%%%%%%%%%%%%%%%%%%%%%%%%%%%%%%%%

\subsection{Domain-Invariant Feature Learning}

% InstanceImageNavでは、ロボットが観察する物体の画像は低画質であるが、提供されるクエリ画像は高画質である。
% したがって、高画質画像と低画質画像の間で、インスタンス単位でドメイン不変な特徴表現を学習することで、タスクの成功率を向上させることができる。
In the InstanceImageNav, the images of objects observed by the robot are low-quality, while the provided query images are high-quality.
Thus, learning instance-wise domain-invariant feature representations between high-quality and low-quality images could improve the success rate of the task.

% 対照学習以外のドメインギャップに対するアプローチ
% 敵対的学習は、領域不変な特徴表現を学習するために利用されてきた。
% このアプローチは、ドメイン分類器の逆勾配をバックプロパゲートすることで、ネットワークがドメイン不変な特徴表現を学習することを可能にする。
% しかし、このアプローチはインスタンス間の識別を見落とす可能性がある。
Adversarial learning has been utilized to learn domain-invariant feature representations~\cite{domain_adversarial_neural_network, adversarial_discriminative_domain_adaptation}.
This approach enables networks to learn domain-invariant feature representations by back-propagating the reverse gradients of the domain classifier.
However, this approach might overlook discrimination between instances.

% しかし、領域不変な特徴表現を得るために対比的学習を用いる手法が提案されている~\cite{contrastive_uda, contrastive_learning_domain_adaptation}.
% 対比学習はネットワークがインスタンス識別的な表現を学習することを可能にするので、対比学習に基づくアプローチはインスタンス単位でドメイン不変な特徴表現を学習することができる。
% そこで、本研究では対比学習に着目する。
% 先行研究~\cite{contrastive_uda, contrastive_learning_domain_adaptation} と大きく異なる点は、本研究では1つのドメインに対して数個のサンプルしか利用できないことである。
However, methods using contrastive learning to obtain domain-invariant feature representations have been proposed~\cite{contrastive_uda, contrastive_learning_domain_adaptation}.
Since contrastive learning enables networks to learn instance discriminative representations~\cite{reverse_engineering}, approaches based on contrastive learning could learn instance-wise domain-invariant feature representations.
Therefore, this study focuses on contrastive learning.
A significant difference from previous work~\cite{contrastive_uda, contrastive_learning_domain_adaptation} is that only a few samples are available in this study for one domain.

%%%%%%%%%%%%%%%%%%%%%%%%%%%%%%%%%%%%%%%%%%%%%%%%%%%%%%%%%%%%%%%%%%%%%%%%%%%%%%%%

\subsection{Motion Deblurring}

% モーションブラーは、センサーの露光時間中にカメラや被写体が動くことで発生し、画像がぼやける。
% そのため、物体探索のために環境中を移動するロボットのセンサから得られる画像は、モーションブラーによって不鮮明になる可能性が高い。
% モーションブラーは、事前に学習された画像認識モデルの認識精度を低下させる。
% そのため、環境内を移動しながら物体を探索するナビゲーションシステムには、画質劣化を除去するメカニズムが必要となる。
Motion blur occurs when the camera or the subject moves during the sensor's exposure time.
Therefore, images obtained from the sensors of a moving robot are likely to be blurred.
Motion blur degrades the recognition accuracy of pre-trained image recognition models~\cite{deblur_gan}.
Therefore, navigation systems that search for objects while in motion require mechanisms to remove image blur.

% デブラーリング法は大きく2つのタイプに分類できる。
% 1つは、デブラーリング問題を最適化問題として定式化する方法である~fast_motion_deblurring, hihg_quality_motion_deblurring, deep_prior}。
% このアプローチでは、勾配降下法を用いて、ぼやけた画像から鮮明な画像へのぼかしカーネルを最適化する。
% もうひとつは学習ベースのアプローチである。
% 学習ベースのアプローチでは、ぼやけた画像と鮮明な画像のペアを用いて、ぼやけた画像から鮮明な画像を復元するようにニューラルネットワークが学習される~\cite{MSSNet, deblur_gan}。
Deblurring methods can be broadly categorized into two types.
One is formulating the deblurring as an optimization problem~\cite{fast_motion_deblurring, hihg_quality_motion_deblurring, deep_prior}.
This approach involves optimizing the blur kernel from the blurred image to the sharp image using gradient descent.
The other is a learning-based approach.
In this approach, neural networks are trained to restore sharp images from blurred images using pairs of blurred and sharp images~\cite{MSSNet, deblur_gan}.
% 
% デブラーリング問題を最適化問題として定式化するアプローチの1つの問題は、デブラーリングに数秒を必要とする高い計算コストである。
% そこで、学習ベースのデブラーリング手法~MSSNet}を利用する。
Formulating deblurring as an optimization problem has a high computational cost.
Therefore, we utilize a learning-based deblurring method~\cite{MSSNet}.

%%%%%%%%%%%%%%%%%%%%%%%%%%%%%%%%%%%%%%%%%%%%%%%%%%%%%%%%%%%%%%%%%%%%%%%%%%%%%%%%

\begin{figure*}[t]
    \centering
    \includegraphics[width=0.98\linewidth]{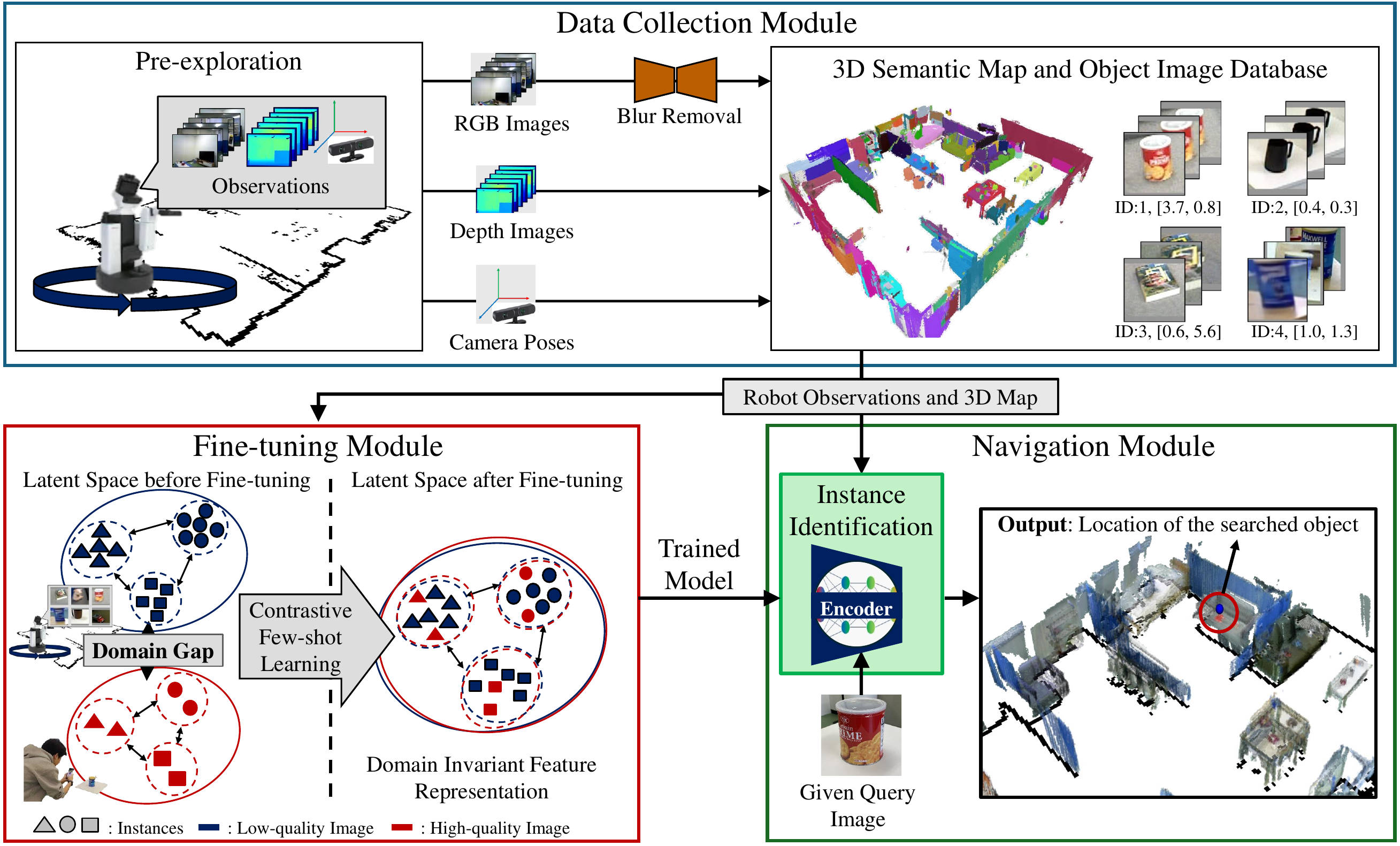}
    \caption{
        % 提案システムの全体構成図。
        % 提案システムは、ロボットが事前に環境を探索して収集したRGBD画像とカメラのポーズ時系列データから物体画像を自動的に収集する。
        % 微調整モジュールは、収集された物体画像を用いて、事前に訓練された画像エンコーダの微調整を行う。
        % ナビゲーションモジュールは、微調整された画像エンコーダを用いて、与えられたクエリ画像と同一のオブジェクトの位置を特定する。
        The overall diagram of the proposed system.
        The proposed system automatically collects object images from RGBD images and camera pose time series data collected by a robot that explored the environment in advance.
        The object image database stores images along with their corresponding ID and position coordinates (e.g., "ID:1, [3.7, 0.8]"). 
        The fine-tuning module fine-tunes the pre-trained image encoder using the collected object images.
        The navigation module identifies the position of the object identical to the given query image using a fine-tuned image encoder.
    }
    \label{fig:overall_system}
\end{figure*}

%%%%%%%%%%%%%%%%%%%%%%%%%%%%%%%%%%%%%%%%%%%%%%%%%%%%%%%%%%%%%%%%%%%%%%%%%%%%%%%%
% Proposed System
%%%%%%%%%%%%%%%%%%%%%%%%%%%%%%%%%%%%%%%%%%%%%%%%%%%%%%%%%%%%%%%%%%%%%%%%%%%%%%%%

\section{Proposed System}
\label{sec:proposed_system}

% Fig.~ref{fig:overall_system}に示すように、提案システムは主に3つのモジュールに分かれている。
% まず、データ収集モジュールは、環境の3次元意味マップを構築し、物体画像を収集する。
% 次に、微調整モジュールが、収集された物体画像とユーザから提供された少数ショットの高画質画像を用いて、対比学習により事前に訓練されたモデルを微調整する。
% コントラスト学習は、「cross-textbf{few-shot cross quality instance-aware adaptation (CrossIA) 」と呼ばれます。｝
% 最後に、ナビゲーションモジュールは、微調整されたモデルと意味マップを活用して、クエリ画像と同じオブジェクトを見つける。
As shown in Fig.~\ref{fig:overall_system}, the proposed system is divided into three main modules.
First, the data collection module constructs the 3D semantic map of the environment and then collects object images.
Second, the fine-tuning module fine-tunes the pre-trained models using the collected object images and few-shot high-quality images provided by a user through contrastive learning.
Our contrastive learning is called {a few-shot cross-quality instance-aware adaptation (CrossIA)}.
Finally, the navigation module leverages the fine-tuned model and the semantic map to locate objects identical to the query image.

%%%%%%%%%%%%%%%%%%%%%%%%%%%%%%%%%%%%%%%%%%%%%%%%%%%%%%%%%%%%%%%%%%%%%%%%%%%%%%%%

\subsection{Data Collection Module}

% データ収集モジュールは、ロボットが3次元空間を移動して収集したRGBD画像とカメラポーズのシーケンスデータから3次元セマンティックマップを構築する。
% 3次元セマンティックマップの構築は、ボクセルベースの3次元マップ～kanechika_3dmap}にインスタンスID情報を投影する。
% ただし、Kanechika~textit{et~al.}がセマンティックセグメンテーションの結果から得られたクラスラベルIDを投影して3次元意味マップを構築するのに対し、我々はFast segment anything (FastSAM)~\cite{fastsam}を利用し、インスタンスID情報を3次元マップに投影して3次元意味マップを構築する。
% 構築された3次元マップがフレーム間で一貫したラベルIDを持つように、3次元意味マップの構築には、Kanechika~textit{et~al.}の手法~kanechika_3dmap}と同様に、Tateno~textit{et~al.}の手法~tateno_3dmap}を採用する。
The data collection module constructs a 3D semantic map from sequence data of RGBD images and camera poses collected by a robot moving in 3D space.
The construction of the 3D semantic map involves projecting instance ID information onto a voxel-based 3D map~\cite{kanechika_3dmap}.
% , similar to the method described by Kanechika et al~\cite{kanechika_3dmap}.
However, while Kanechika~\textit{et~al.} construct a 3D semantic map by projecting the class label IDs obtained from the semantic segmentation results, we utilize fast segment anything (FastSAM)~\cite{fastsam} to construct the 3D semantic map by projecting instance ID information onto the 3D map.
To ensure that the constructed 3D map has consistent label IDs from frame to frame, we employ 3D semantic mapping method~\cite{tateno_3dmap} similar to Kanechika~\textit{et~al.}'s approach~\cite{kanechika_3dmap}.

% まず、ロボットが環境を探索することによって収集されたRGB画像は、デブ ラリングのためにマルチスケールステージネットワーク～multi-scite{MSSNet}に入力される。
% 次に、FastSAM~fastsam}に入力し、画像セグメンテーションを行う。
% そして、セグメンテーション結果、深度画像、カメラポーズを用いて3次元意味マップを構築する。
First, RGB images collected by the robot exploring the environment are entered into the multiscale stage network~\cite{MSSNet} for deblurring.
Next, FastSAM segments images element by element.
% they are inputted into FastSAM~\cite{fastsam} for image segmentation.
The 3D semantic map is then constructed using the segmentation results, depth images, and camera poses.

% 次に、このモジュールは、構築された3Dセマンティックマップを使用してオブジェクト画像を収集する。
% この過程で、2Dセグメンテーションはレイトレーシングによって3Dセマンティックマップから画像をマスクする。
% レイトレーシングは、カメラから奥行き方向に擬似光線を送り、3Dマップ上の最初の衝突のインスタンスIDをキャプチャすることにより、任意のカメラポーズからマスク画像を生成する。
% この処理により、同一インスタンスの画像に対する擬似ラベルの一貫した生成が保証される。
% その後、生成されたマスク画像をバウンディングボックス（BBox）に変換し、変換されたBBoxの領域をRGB画像から抽出してオブジェクト画像を収集する。
% % Next, this module collects object images using a constructed 3D semantic map.
% During this process, 2D segmentation masks images from the 3D semantic map by ray tracing.
% Ray tracing generates a mask image from an arbitrary camera pose by sending pseudo-rays in the depth direction from the camera and capturing the instance ID of the first collision on the 3D map.
% This process ensures the consistent generation of pseudo-labels for images of the same instance.
% Subsequently, the generated mask images are transformed into bounding boxes (BBoxes), and the regions of the transformed BBoxes are extracted from the RGB images to collect object images.

% このプロセスでは、レイトレーシングと呼ばれる手法を用いて、3Dセマンティックマップから2Dセグメンテーションマスクが生成される。レイトレーシングは、任意のカメラポーズから奥行き方向に擬似光線を投影することで機能する。
% これらの光線は3Dマップと相互作用し、光線によって検出された最初の衝突のインスタンスIDがキャプチャされる。その結果、3Dマップの分割された領域に対応するマスク画像が作成される。
% この方法により、異なるフレーム間で同じインスタンスの画像に対して一貫して擬似ラベルが生成されることが保証される。マスク画像が作成されると、それらはバウンディングボックス（BBox）に変換される。
% これらのバウンディングボックスは、RGB画像から特定の領域を抽出するために使用され、セグメント化された領域に対応するオブジェクト画像の収集を可能にする。
During this process, 2D segmentation masks are generated from the 3D semantic map using a method called ray tracing. 
% Ray tracing works by projecting pseudo-rays from an arbitrary camera pose in the depth direction. 
These rays interact with the 3D map, and the instance ID of the first collision detected by the rays is captured. This results in the creation of a mask image that corresponds to the segmented area of the 3D map.
This method ensures that pseudo-labels are consistently generated for images of the same instance across different frames. Once the mask images are created, they are transformed into bounding boxes (BBoxes). 
These bounding boxes are used to extract specific regions from the RGB images, allowing for the collection of object images that correspond to the segmented areas.

%%%%%%%%%%%%%%%%%%%%%%%%%%%%%%%%%%%%%%%%%%%%%%%%%%%%%%%%%%%%%%%%%%%%%%%%%%%%%%%%

\subsection{Fine-tuning Module}

% このモジュールは、ロボットが観察する低画質画像と、ユーザが提供する数ショットの高画質画像との対比タスクによって、事前に訓練された画像エンコーダを微調整する。
% SimSiamは否定的自由対照学習法であり、少ないバッチサイズで学習できる。
% これに対して、コントラスト学習法では、大きなバッチサイズ~\cite{simclr}で否定ペア学習を行う必要がある。
% そこで、SimSiamを利用して微調整を行う。
This module fine-tunes a pre-trained image encoder by the contrastive learning between low-quality images observed by the robot and few-shot high-quality images provided by the user.
SimSiam, a negative free contrastive learning method, can learn with a small batch size~\cite{simsiam}.
In contrast, contrastive learning methods require negative pair learning with a large batch size~\cite{simclr}.
Therefore, we utilize SimSiam for fine-tuning.

% また、トレーニングの際には、ロボットに位置を特定させたい対象物を携帯端末で数枚撮影し、ロボットに提供する必要がある。
% これは、ユーザがロボットに指示する必要があるが、数枚の画像であるため、ユーザにとって大きな負担にはならない。
% 本研究では、システムが各インスタンスに対して最大5枚の高品質画像を持つと仮定し、それに従って実験を評価する。
Additionally, for training, the user must capture a few images of the objects they want the robot to locate using a mobile device and provide these images to the robot. 
% Although this requires the user to interact with the robot, t
The process is relatively simple and involves only a small number of images, so the user is not significantly burdened.
% it is not a significant burden on the user.
% In this study, we assume that the system has at most five high-quality images for each instance and evaluate the experiments accordingly.

\begin{figure}[t]
    \centering
    \includegraphics[width=0.96\linewidth]{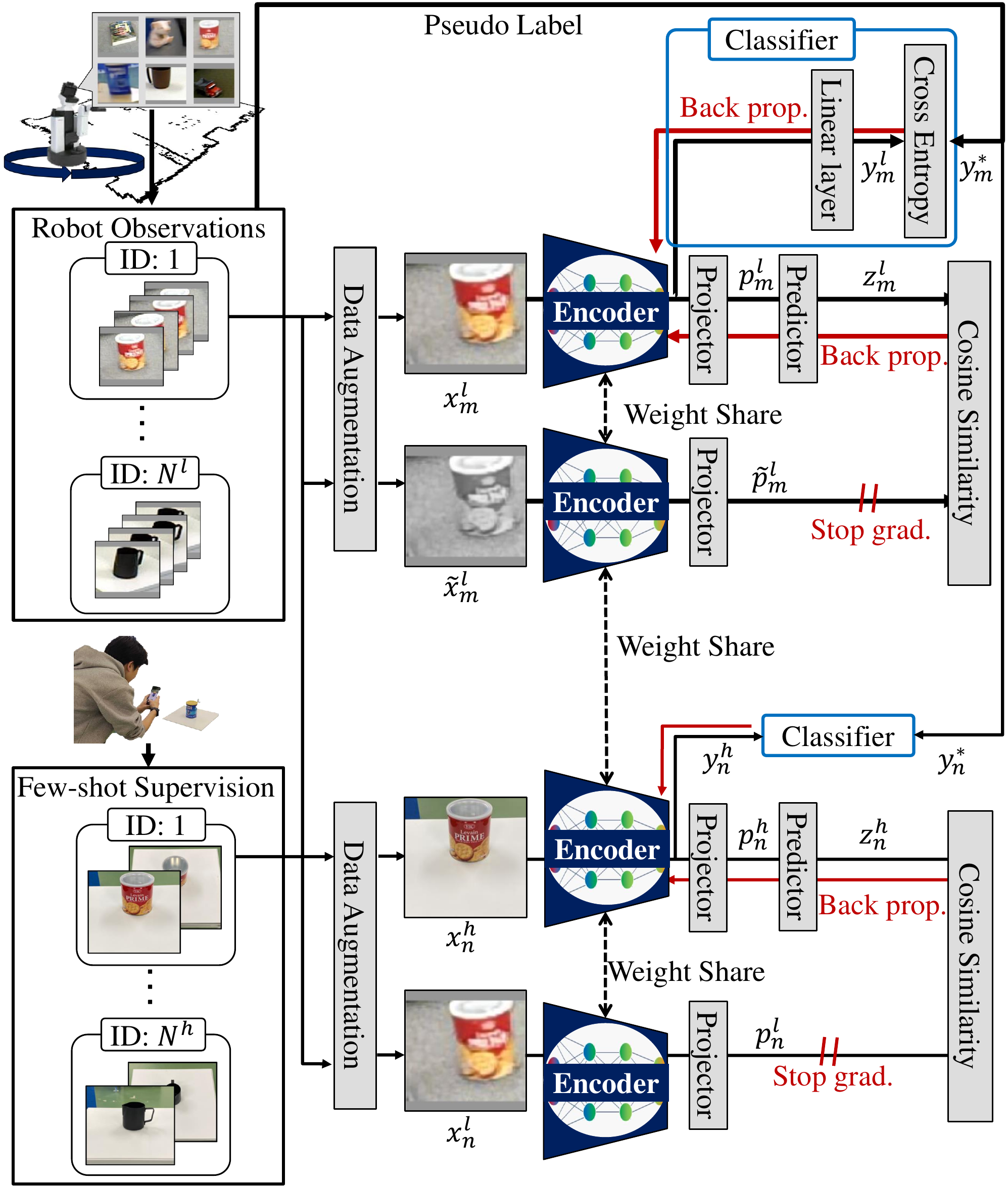}
    \caption{
        The fine-tuning by few-shot cross-quality instance-aware adaptation.
        Two contrasting images are swapped and processed in the same way.
    }
    \label{fig:fine_tuning_module}
    \vspace{-6pt}
\end{figure}

% 画像エンコーダの微調整には、高画質画像と低画質画像の対比学習と、ロボットが観察した画像間の対比学習が含まれる。
Fine-tuning the image encoder involves contrastive learning between high-quality and low-quality images, and the contrastive learning between images the robot observed, as shown in Fig.~\ref{fig:fine_tuning_module}.
% さらに、fine-tuningによって、事前に訓練された画像エンコーダに対する対比学習と線形分類器の損失を最小化することで、同じラベルの画像間の特徴ベクトルの分散が減少する〜cite{contrastive_fine_tuning}。
% この特性は画像分類タスクの精度向上につながる。
% そこで、図～ref{fig:fine_tuning_module}に示すように、SimSiamに線形分類器を追加して、画像エンコーダの微調整も行う。
% 線形分類器の損失計算に必要なインスタンスのラベルは、データ収集モジュールによって自動生成された擬似ラベルを用いて得られる。
% 従って、微調整時の損失関数は以下のように表現できる:
Furthermore, minimizing the losses associated with contrastive learning and the linear classifier during fine-tuning of a pre-trained image encoder reduces the variance of feature vectors between images with the same label~\cite{sakaguchi_iros2024, contrastive_fine_tuning}.
% This characteristic leads to an improvement in the accuracy of image classification tasks.
Consequently, we perform fine-tuning of the image encoder by adding a linear classifier to SimSiam, as illustrated in Fig.~\ref{fig:fine_tuning_module}.
The labels required for calculating the linear classifier's loss are obtained using pseudo-labels automatically generated by the data collection module.
Thus, the loss function during fine-tuning can be expressed as follows:
\begin{equation}
    \label{eq:fine_tuning}
    \begin{split}
        \mathcal{L} = \sum_{m=1}^{M}\mathcal{L}^{\textrm{robot}}_{m} + \sum_{n=1}^{N}\mathcal{L}^{\textrm{cross}}_{n},
    \end{split}
\end{equation}
where $M$ and $N$ represent the number of pairs of low-quality images and the number of pairs of high-quality and low-quality images, for the same instance in a mini-batch, respectively.
% where $N$ denotes the number of instances. 
% $N^{\textrm{cross}}_n$ and $N^{\textrm{robot}}_n$ denote the number of high-quality images and the number of low-quality images, in the instance ID $n$, respectively.
Loss functions $\mathcal{L}^{\textrm{robot}}_{m}$ and $\mathcal{L}^{\textrm{cross}}_{n}$ are calculated as follows:
\begin{equation}
    \label{eq:in_domain_contrast}
    \begin{split}
        \mathcal{L}^{\textrm{robot}}_{m} &= -\frac{1}{2}\{\text{CosSim}(p_{m}^{l}, \tilde{z}_{m}^{l}) + \text{CosSim}({\tilde{p}}_{m}^{l}, {{z}}_m^{l} )\} \\
        &\quad + \frac{1}{2} \{ \text{CE}( {y}_{m}^{l}, y^{\ast}_{m} ) +\text{CE}( {\tilde{y}}_{m}^{l}, y^{\ast}_{m}) \},
    \end{split}
\end{equation}
\begin{equation}
    \label{eq:cross_domain_contrast}
    \begin{split}
        \mathcal{L}^{\textrm{cross}}_{n} &= -\frac{1}{2}\{\textrm{CosSim}(p_n^l, z_n^h) + \textrm{CosSim}(p_n^h, z_n^l)\} \\
        &\quad + \frac{1}{2}\{\textrm{CE}({y}_{n}^{h}, y^{\ast}_{n}) +\textrm{CE}({y}_{n}^l, y^{\ast}_{n})\}, %\\
    \end{split}
\end{equation}
% ここでCE()とCosSim()はクロスエントロピーと余弦類似度を表す。
% p_i^l$と$z_i^{l}$はそれぞれ、ロボットが観測した物体画像$x_i^l$に対するSimSiamの投影器と予測器の出力である。
% p_i^{prime}}^{l}$と${z_i^{prime}}^{l}$は、SimSiamの投影器と物体画像の予測器の出力${x_i^{prime}}^{l}$である。
% y_{i, pred}^l$は$x_i^l$が分類されたときの予測結果である。
% p_i^h$と$z_i^h$はそれぞれ、高画質画像$x_i^h$がSimSiamに入力されたときの投影器と予測器の出力であり、$y_{i, pred}^h$は$x_i^l$が分類されたときの予測結果である。h$は$x_i^h$が分類されたときの予測結果である。
where CosSim() and CE() denote cosine similarity and cross-entropy.
Variables $p$ and $z$ with a certain subscript are the outputs of SimSiam's projector and predictor, respectively.
The superscripts $l$ and $h$ denote that they are derived from low- and high-quality images.
The same applies to the variables with tildes.
Here, $x$ and ${\tilde{x}}$ are different images for the same object generated by data augmentation.
A variable $y$ is the prediction ID when the image $x$ is classified.
$y^{\ast}$ is the pseudo-label of the instance obtained from the data collection module.

% $p^l$ and $z^l$ are the outputs of SimSiam's projector and predictor of the low-quality image $x^l$ observed by the robot, respectively.
% $p^h$ and $z^h$ are the outputs of SimSiam's projector and predictor of the high-quality image $x^h$.
% ${\tilde{p}}^{l}$ and ${\tilde{z}}^{l}$ are the outputs of SimSiam's projector and predictor of the low-quality image ${\tilde{x}}^{l}$. 
% $x^{l}$ and ${\tilde{x}}^{l}$ are different images of the same object generated by data augmentation.
% ${y}_{i}^l$ is the prediction result when $x_i^l$ is classified.
% $p_i^h$ and $z_i^h$ are the outputs of the projector and predictor, respectively, when the high-quality image $x_i^h$ is input to SimSiam, and ${y}_{i}^h$ is the prediction result when $x_i^h$ is classified.
% $y^{\ast}$ is the instance ID obtained from the data collection module.

% 本研究では、${(x_i^l, x_i^h, y_{i, true}) \}_{i =1}^{N_c}$を高画質画像、低画質画像とそれらのラベルの集合、${(x_i^l, y_{i,true}) \}_{i=1}^{N_r}$ を低画質画像の画像とそれらのラベルの集合とする。
% $\{(x_i^l, x_i^h, y^{\ast}_{i})\}_{i =1}^{N^{\textrm{cross}}_n}$ is the set of high-quality image, low-quality image and their label, and $\{(x_i^l, y^{\ast}_{i})\}_{i=1}^{N^{\textrm{robot}}_n}$ is the set of images and their labels of low-quality image.

%%%%%%%%%%%%%%%%%%%%%%%%%%%%%%%%%%%%%%%%%%%%%%%%%%%%%%%%%%%%%%%%%%%%%%%%%%%%%%%%

\subsection{Navigation Module}

% 図～ref{fig:overall_system}に示すように、ナビゲーションモジュールは与えられたクエリ画像と同じオブジェクトの位置を特定する。
% これは、データ収集モジュールによって収集されたオブジェクトの画像、3Dセマンティックマップ、および微調整された画像エンコーダから抽出された特徴ベクトルを活用することによって行われる。
As shown in Fig.~\ref{fig:overall_system}, the navigation module finds the object that matches the given query image. 
It uses images collected by the data collection module, the 3D semantic map, and feature vectors from the fine-tuned image encoder.

% As shown in Fig.~\ref{fig:overall_system}, the navigation module locates the object identical to the given query image.
% This is conducted by leveraging images of objects collected by the data collection module, the 3D semantic map, and feature vectors extracted from the fine-tuned image encoder.

% まず、微調整された画像エンコーダを利用して、データ収集モジュールによって収集された物体画像とクエリ画像の特徴ベクトル$z_{\textrm{obs}}$と$z_{\textrm{query}}$をそれぞれ計算する。
% これらの特徴ベクトルを用いて、以下のように示す余弦類似度を用いて、クエリ画像と観測物体画像との類似度を計算する：
First, the fine-tuned image encoder is utilized to compute the feature vectors $z_{\textrm{obs}}$ and $z_{\textrm{query}}$ for the object images collected by the data collection module and query image, respectively.
Using these feature vectors, the similarity between the query and observed object images is calculated using the cosine similarity shown as follows:
\begin{equation}
    \label{eq:cosine_simirality}
    \text{CosSim}(z_{\textrm{query}}, z_{\textrm{obs}}) = \frac{z_{\textrm{query}}}{\|z_{\textrm{query}}\|_2}\cdot\frac{z_{\textrm{obs}}}{\|z_{\textrm{obs}}\|_2}.
\end{equation}
% その後、クエリ画像との類似度が最も高いインスタンスが特定される。
Subsequently, the instance with the highest similarity to the query image is identified.

% 次に、モジュールはクエリ画像に最も似ている3Dマップ上のオブジェクトの重心座標を特定し、ロボットはそれに近づくために移動する。
% オブジェクトの重心座標は、識別されたインスタンスと同じIDを持つボクセルの3D座標の平均を計算することによって決定される。
% 物体の重心座標を目標点として決定する際、地図上の占有位置を目標点とした場合、ナビゲーションに失敗する可能性がある。
% したがって、この問題を克服するために、モジュールは、オブジェクトのセントロイドから半径1.0m以内の占有されていない領域をナビゲーションのターゲットポイントとして選択する。
% この1.0mという値は、Krantz~textit{et~al.}のInstanceImageNavタスクの成功基準の定義に基づいて選択されたもので、目標物体とロボットの間の距離は1.0m以下と定義されている。
Next, the module identifies the centroid coordinates of the object on the 3D map that best matches the query image, and the robot moves towards it.
The centroid is calculated by averaging the 3D coordinates of the voxels with the same ID as the identified object. However, navigation can fail if the centroid's location overlaps with an occupied position on the map. 
To address this, the module selects an unoccupied area within a 1.0 m radius from the object's centroid as the navigation target. 
This distance is based on Krantz~\textit{et~al.}'s definition of the success criteria for the InstanceImageNav task, where the distance between the target object and the robot is defined as 1.0 m or less~\cite{instanceimagenav}.

% Next, the module identifies the centroid coordinates of the object on the 3D map that is most similar to the query image, and the robot moves to close it.
% The centroid coordinates of the object are determined by computing the average of the 3D coordinates of the voxels with the same ID as the identified instance.
% Upon determining the centroid coordinates of the object as the target point, navigation may fail if the occupied position on the map is designated as the target point.
% Therefore, to overcome this issue, the module selects unoccupied regions within a radius of 1.0 m from the object's centroid as the target point for navigation.
% This value is chosen based on Krantz~\textit{et~al.}'s definition of success criteria for the InstanceImageNav task, where the distance between the target object and the robot is defined as 1.0 m or less~\cite{instanceimagenav}.

%%%%%%%%%%%%%%%%%%%%%%%%%%%%%%%%%%%%%%%%%%%%%%%%%%%%%%%%%%%%%%%%%%%%%%%%%%%%%%%%
% Experiments
%%%%%%%%%%%%%%%%%%%%%%%%%%%%%%%%%%%%%%%%%%%%%%%%%%%%%%%%%%%%%%%%%%%%%%%%%%%%%%%%

\section{Experiments}
\label{sec:experiments}

% 提案システムが、ナビゲーションモジュールを用いて観測画像からクエリ画像と同一の物体を正確に特定できれば、目的物体への移動が可能となる。
% そこで、本実験では、観察画像からクエリ画像と同一の物体を特定することで、提案システムの評価を行う。
% 本実験の目的は2つある。
% \begin{列挙｝
%     \項目 環境探索時にぼかしを導入することで、タスクの成功率が向上するか検証します。
%     \itemロボットが観察した画像とユーザが提供した数ショットの高画質画像との対比学習により事前学習したモデルを微調整することで、タスクの成功率が向上するかを検証します。
% \終了
If our proposed system can accurately identify the object that matches the query image within the observed images using the navigation module, then moving to the target object should be feasible. 
Therefore, in this experiment, we evaluate the system by checking how well it identifies the object that matches the query image from the observed images. 
This experiment has two main objectives:
\begin{enumerate}
    \item To determine if introducing deblurring during exploration in the environment can improve the success rates.
    \item To assess whether fine-tuning a pre-trained model using contrastive learning with both images observed by the robot and a few high-quality images provided by the user can enhance success rates.
    % We verify whether fine-tuning a pre-trained model using contrastive learning with images observed by the robot and a few-shot high-quality images provided by the user can improve the success rates.
\end{enumerate}
% ファインチューニング時に高画質画像を数枚提供する効果を検証するため、各対象に対して提供する画像の枚数が少ないほどタスク成功率が向上し、どのように低下するかを調べた。
In addition, to evaluate the impact of using a few high-quality images during fine-tuning, we examined how task success rates improve and how they decline as fewer images are provided for each object. 

The system was developed using a software development environment~\cite{el_hafi_software_2022}.

%%%%%%%%%%%%%%%%%%%%%%%%%%%%%%%%%%%%%%%%%%%%%%%%%%%%%%%%%%%%%%%%%%%%%%%%%%%%%%%%

\subsection{Comparison Methods}

% この実験では、InstanceImageNavの先行研究で利用されているSuperGlue～superglue、SimSiam～simsiamで事前に学習された画像エンコーダ、提案するCrossIAを比較する。
% SimSiamは、オープンデータセット〜cite{scannet, imagenet}で事前に訓練されたモデルを使用。
% SimView~\cite{sakaguchi_iros2024}は、式３のみ。アブレーションとしても機能する。
We compare SuperGlue~\cite{superglue}, SimSiam~\cite{simsiam}, SimView~\cite{sakaguchi_iros2024}, and the proposed CrossIA.
SuperGlue was utilized in previous work on InstanceImageNav~\cite{modular_instancenav, goat}.
SimSiam used a pre-trained model on open dataset~\cite{scannet, imagenet}.
SimView is equivalent to the case where only Eq.~\eqref{eq:in_domain_contrast} is employed and also serves as an ablation study for the proposed method.
% The baselines are the condition where models pre-trained on open dataset~\cite{scannet, imagenet} are utilized without utilizing a deblurring method to identify the object identical to the query image.
% 
% また、ドメインラベルを用いた対比学習と敵対学習を組み合わせることで、ドメイン不変特徴表現を学習する手法を評価した～domain_adversarial_neural_network}。
% 学習時に提供される高品質な画像の量が少ない場合でも、ドメイン不変な特徴表現を学習することができる。
In addition, we evaluated a method to learn domain-invariant feature representations by combining contrastive learning and adversarial learning with domain labels~\cite{domain_adversarial_neural_network}.
Domain-invariant feature representation can be learned even when the amount of high-quality images provided during training is small.

% \begin{figure}[t]
%     \centering
%     \begin{tabular}{cc}
%       \begin{minipage}[b]{0.47\linewidth}
%         \centering
%         \includegraphics[keepaspectratio, width=0.9\linewidth]{figures/used/hm3d_rgb_mesh_v1.pdf}
%         \subcaption{RGB 3D mesh}
%       \end{minipage}
%       % 
%       \begin{minipage}[b]{0.47\linewidth}
%         \centering
%         \includegraphics[keepaspectratio, width=0.9\linewidth]{figures/used/hm3d_semantic_mesh_v1.pdf}
%         \subcaption{Semantic 3D mesh}
%       \end{minipage}
%       % 
%     \end{tabular}
%   \caption{Example 3D mesh data which is included in HM3D.}
%   \label{fig:hm3d_example_3dmesh}
% \end{figure}

%%%%%%%%%%%%%%%%%%%%%%%%%%%%%%%%%%%%%%%%%%%%%%%%%%%%%%%%%%%%%%%%%%%%%%%%%%%%%%%%

\subsection{Conditions and Dataset}

\begin{figure}[t]
    \centering
    \includegraphics[width=0.88\linewidth]{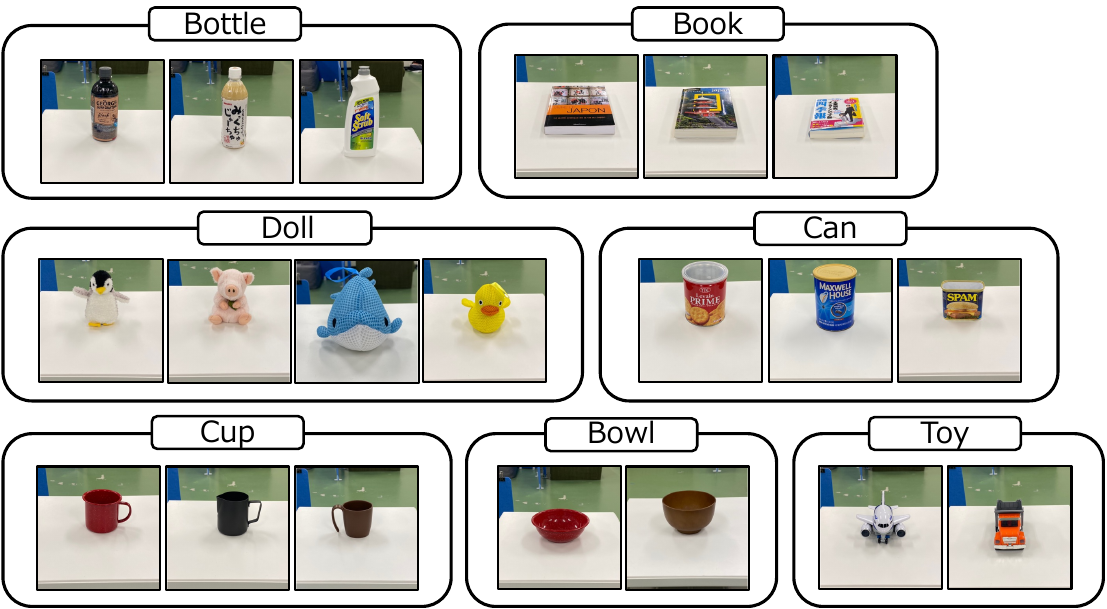}
    \caption{
        Instances targeted for search in this experiment.
    }
    \label{fig:target_instances}
\end{figure}

% 本実験で収集したロボット観察データは、トヨタ自動車株式会社製のHSR（Human Support Robot）～HSRを使用した。
% また、HSRに取り付けられているRGBDセンサはAsus Xtion Pro Liveである。
% このカメラは、解像度640$×480$、30fpsのRGB画像を撮影することができる。
% RGBDセンサーは地上から約1mの高さに取り付けられている。
% % 
% データ収集中、ロボットは特定の物体の前で停止することなく環境全体を移動し、74[$text{m}^{2}$]エリア全体を2分間でスキャンし、3Dセマンティックマップを構築した~\cite{tatenoo_3dmap}。
% この探索では606枚のRGBD画像を収集した。
% 構築されたデータセットには145のインスタンスが含まれ、2011の画像がロボットによって収集された。
The robot observation data collected in this experiment were obtained using the Human Support Robot (HSR)~\cite{hsr} built by Toyota Motor Corporation.
Additionally, the RGBD sensor attached to the HSR is the Asus Xtion Pro Live.
This camera can capture RGB images with a resolution of 640$\times$480 at 30 fps.
The RGBD sensor is mounted about 1 m above the ground.
During data collection, the robot moved throughout the environment without stopping in front of specific objects, and the entire 74 $\text{m}^{2}$ area was scanned in 2 min to construct a 3D semantic map~\cite{tateno_3dmap}.
This exploration collected 606 RGBD images.
The constructed dataset contains 145 instances, and 2011 images were collected by the robot.

% 各インスタンスの高品質なクエリ画像は、iPhone 11 Proのカメラで撮影された。
The high-quality images for each instance were captured using the camera on an iPhone 11 Pro.
% 
%%%トレーニング用のfew-shot画像について
% Fig.~ref{fig:target_instances}の各インスタンスに5ショットの高画質画像を提供し、微調整を行った。
% また、アブレーションスタディとして、1ショット、3ショット、5ショットを比較した。
We provided five high-quality images for each instance for fine-tuning. 
They are all different arrangements and backgrounds in the environment.
Additionally, we conducted ablation studies comparing one, three, and five images. 
% To evaluate the impact of using a few high-quality images during fine-tuning, we examined how task success rates improve and how they decline as fewer images are provided for each object. 

% ロボットは、図～ref{fig:target_instances}に示すように、〚ボトル〛、〛カップ〛、〛缶〛、〛ボウル〛、〛人形〛、〛玩具〛のクラスに属するインスタンスを探し出す。
The robot locates 20 instances from the classes \texttt{Bottle}, \texttt{Book}, \texttt{Cup}, \texttt{Can}, \texttt{Bowl}, \texttt{Doll}, and \texttt{Toy}, as shown in Fig.~\ref{fig:target_instances}.
% 
% クエリ画像を撮影する際、撮影者は40[cm]から撮影した。
% そこで、本実験では、クエリ画像を撮影する際に、撮影者が立ったりしゃがんだりしながら8方向から撮影し、各インスタンスに対して32種類のクエリ画像を用意した。
%%%評価用のクエリについて
The query images were taken from distances of 40 cm.
When capturing the images, the photographer took pictures from eight different angles, both standing and crouching, resulting in 32 images for each instance.
Note that the high-quality images used for training and testing are different.

% The high-quality images for each instance were taken using the camera of an iPhone 11 Pro.
% When taking the images, the photographer took the pictures from 40 and 50 [cm].
% Moreover, in studies addressing the InstanceImageNav task in the real world, only one image is captured for each instance.
% However, q
% Query images taken by users may be captured from various angles.
% When capturing the images, the photographer captured pictures from eight different directions, standing and crouching, to prepare 32 types of images for each instance.

%%%%%%%%%%%%%%%%%%%%%%%%%%%%%%%%%%%%%%%%%%%%%%%%%%%%%%%%%%%%%%%%%%%%%%%%%%%%%%%%

\subsection{Training Set-up}
% The workstation is equipped with an AMD Ryzen Threadripper3 3960X CPU, 64GB RAM, and two Nvidia RTX A6000 GPUs.
The PCs are equipped with an Intel Core i9-9900K CPU, 64GB RAM, and an Nvidia RTX 2080 GPU for training, excluding SimView, and an Intel Core i9-14900KF CPU, 64GB RAM, and an Nvidia RTX 4090 GPU for training with SimView and all evaluation.
% CPU ：Intel Core i9-14900KF
% G/PU ：Nvidia RTX 4090
% RAM ：64 GB.
% 学習条件間で統一したハイパーパラメータの設定を示す．
% いずれの条件においてもオプティマイザには，Stochastic Gradient Descent (SGD)を用いた．
% 学習率については，Cosine Scheduler~\cite{cosine-scheduler}を用いて学習が進むにつれ減衰させるようにした．
% また，学習時に用いたデータ拡張の種類は，crop and resize, color jitter, grayscale, and horizontal filpの4種類の手法を利用している．
The optimizer used for all conditions was stochastic gradient descent.
The batch size was set to 256 and the learning rate was 0.07. 
The training was conducted for 1000 epochs. 
These hyperparameters were chosen to ensure consistency and comparability in our experiments.
Four types of data augmentation were used during the training: crop and resize, color jitters, grayscale, and horizontal flip.

% In the training process, we unified several hyperparameters across different training conditions. 
% We used a weight decay of $1.5 \times 10^{-6}$ and a momentum of 0.9. 
% The training was conducted for a total of 1000 epochs. 
% The learning rate was allowed to decay as the training progressed using the Cosine Scheduler. 
%~\cite{cosine_scheduler}

%%%%%%%%%%%%%%%%%%%%%%%%%%%%%%%%%%%%%%%%%%%%%%%%%%%%%%%%%%%%%%%%%%%%%%%%%%%%%%%%

\subsection{Evaluation Metrics}

% タスクの評価指標には、成功率(SR)、平均相互順位(MRR)~eqite{mrr}、平均順位 (MR)が含まれる。
% SRは、ロボットが観測した物体画像の中で、クエリ画像と同一の物体を上位に特定する精度を評価する指標である。
% MRRは、クエリ画像と同一の物体を上位のランクでどれだけ識別できるかを評価する指標である。
The evaluation metrics for the task include the success rate (SR), the mean reciprocal rank (MRR)~\cite{mrr}, and the mean rank (MR).
% calculated using equations Eq.~\eqref{eq:success_rate} and Eq.~\eqref{eq:mean_reciprocal_rank}.
% SR is used to evaluate the accuracy of identifying the object identical to the query image as the top-ranked item among the images of objects observed by the robot.
SR evaluates how accurately the object matching the query image is identified as the most similar object among the images observed by the robot.
MRR measures how well the object matching the query image is identified at higher ranks.
% MRR is a metric that evaluates how well the object identical to the query image can be identified at higher ranks.
Thus, they can be expressed as follows:
\begin{equation}
    % \label{eq:success_rate}
    \text{SR} = \frac{1}{N}\sum^{N}_{n=1}s_n ,
    \qquad
% \end{equation}
% \begin{equation}
    % \label{eq:mean_reciprocal_rank}
    \text{MRR} = \frac{1}{N}\sum^{N}_{n=1}\frac{1}{k_n},
    \label{eq:success_rate}
\end{equation}
where $N$ is the number of trials and $N=640$ (i.e. 20 instances $\times$ 32 images). for this study. 
% ここで$N$は試行回数であり、本研究では合計640枚のクエリ画像を収集したので$N=640$である。
% 式~eqref{eq:success_rate}において、$s_n$は各インスタンスでタスクが成功したかどうかを示す二値変数で、クエリ画像タスクが成功した場合は$s_n=1$、そうでない場合は$s_n=0$である。
% k_n$は、クエリ画像と同じオブジェクトが検索されたランクである。
% where $N$ is the number of trials, which in this study $N=640$ since a total of 640 query images were collected. 
In Eqs.~\eqref{eq:success_rate}, $s_n$ represents a binary variable indicating whether the task was successful for each instance, where $s_n=1$ if the query image task was successful and $s_n=0$ otherwise.
$k_n$ is the rank at which the same object as the query image is retrieved.

% MRはMRRの逆数であり、タスクの全試行を通じて、クエリ画像と同じオブジェクトが平均して検索された順位を示す指標である。
MR is the reciprocal of MRR and indicates the average rank at which the object matching the query image was found across all trials of the task.
% MR is the reciprocal of MRR and is a metric that indicates the rank in which the same object as the query image was searched on average across all trials of the task.

%%%%%%%%%%%%%%%%%%%%%%%%%%%%%%%%%%%%%%%%%%%%%%%%%%%%%%%%%%%%%%%%%%%%%%%%%%%%%%%%
% Results
%%%%%%%%%%%%%%%%%%%%%%%%%%%%%%%%%%%%%%%%%%%%%%%%%%%%%%%%%%%%%%%%%%%%%%%%%%%%%%%%

\section{Results}
\label{sec:results}

\begin{table}[t]
    \begin{center}
        \caption{% Results of Fine-tuned SimSiam and Pre-trained Models
                 Score Results of Comparison Methods
                }
        \label{tab:all_results}
        \begin{tabularx}{1.0\linewidth}{|l|c|CCC|} 
            \hline
            \textbf{Methods} & \textbf{Deblurring} & \textbf{SR $\uparrow$} & \textbf{MRR $\uparrow$} & \textbf{MR $\downarrow$} \\
            \hline
            \hline
            SuperGlue~\cite{superglue} & --- & $0.275$ & $0.365$ & $2.73$ \\
            SuperGlue~\cite{superglue} & \checkmark & $0.281$ & $0.370$ & $2.70$ \\
            SimSiam~\cite{simsiam} & --- & $0.293$ & $0.413$ & $2.41$ \\
            SimSiam~\cite{simsiam} & \checkmark & ${0.290}$ & ${0.416}$ & ${2.40}$ \\
            % \hline
            SimView~\cite{sakaguchi_iros2024} & ---        & $0.066$ & $0.115$ & $8.64$ \\
            SimView~\cite{sakaguchi_iros2024} & \checkmark & $0.034$ & $0.097$ & $10.3$ \\
            % SimView~\cite{sakaguchi_iros2024} & ---        & $0.0656$ & $0.115$ & $8.64$ \\
            % SimView~\cite{sakaguchi_iros2024} & \checkmark & $0.0343$ & $0.0966$ & $10.3$ \\
            CrossIA (Ours)  & \checkmark & $\underline{\bm{0.751}}$ & $\underline{{0.812}}$ & $\underline{{1.24}}$ \\ 
            CrossIA with~\cite{domain_adversarial_neural_network} & \checkmark & $\underline{0.753}$ & $\underline{\bm{0.820}}$ & $\underline{\bm{1.21}}$ \\ 
            \hline
            % \multicolumn{5}{p{0.9\linewidth}}{
            %     \vspace{0.1pt}
            %     Scores for all evaluation metrics.
            %     MR represents the average rank of similarity between the query image and the same object.
            % }
        \end{tabularx}
    \end{center}
\end{table}

\begin{table}[t]
    \begin{center}
        \caption{Scores of fine-tuned SimSiam based on few-shot conditions}
        \label{tab:ablation_all_results}
        \begin{tabularx}{1.0\linewidth}{|L|CCC|}
            \hline
            \textbf{Methods} & \textbf{SR $\uparrow$} & \textbf{MRR $\uparrow$} & \textbf{MR $\downarrow$} \\
            \hline
            \hline
            One-shot & {0.421} (0.587) & {0.548} (0.696) & {1.82} (1.43) \\
            % \hline
            Three-shot & {0.646} (0.690) & {0.752} (0.763) & {1.32} (1.31) \\
            % \hline
            Five-shot & {0.751} (0.753) & {0.812} (0.820) & {1.24} (1.21) \\
            \hline
            \multicolumn{4}{p{0.9\linewidth}}{
                \vspace{0.1pt}
                % 内の数字は、敵対的学習を加えた結果である。
                The numbers in parentheses are the results with the addition of adversarial learning~\cite{domain_adversarial_neural_network}.
                % The reciprocal of MRR represents, on average across all trials, at what position the target instance ID is searched for.
            }
        \end{tabularx}
    \end{center}
\end{table}

\begin{figure*}
    \centering
    \includegraphics[width=0.86\linewidth]{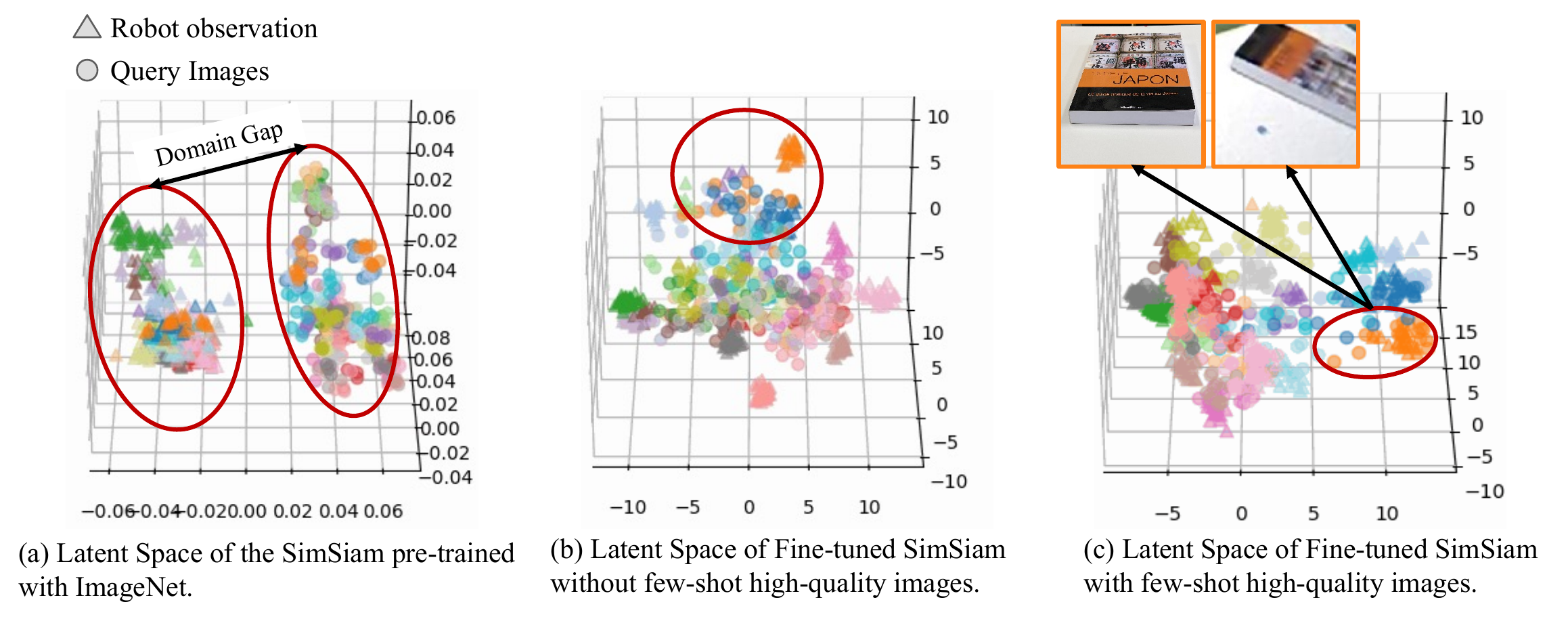}
    \caption{
        % 異なる学習条件におけるSimSiamの潜在空間。
        % (a）異なるドメイン間のデータは空間的に分離されている。
        % (b)赤い楕円で示した領域のように、同じインスタンスの低画質画像のみが近接している。
        % (c)同じインスタンスの異なるドメインからの画像は、潜在空間において近くなる。
        Latent spaces of the image encoders for different training conditions.
        (a) SimSiam: Data between different domains are spatially separated.
        (b) SimView: Only low-quality images of the same instance are closer together, as shown in the region marked by the red ellipse.
        (c) CrossIA: Images from different domains of the same instance move closer in the latent space.
    }
    \label{fig:latent_spaces}
\end{figure*}

\begin{figure*}
    \centering
    \includegraphics[width=0.82\linewidth]{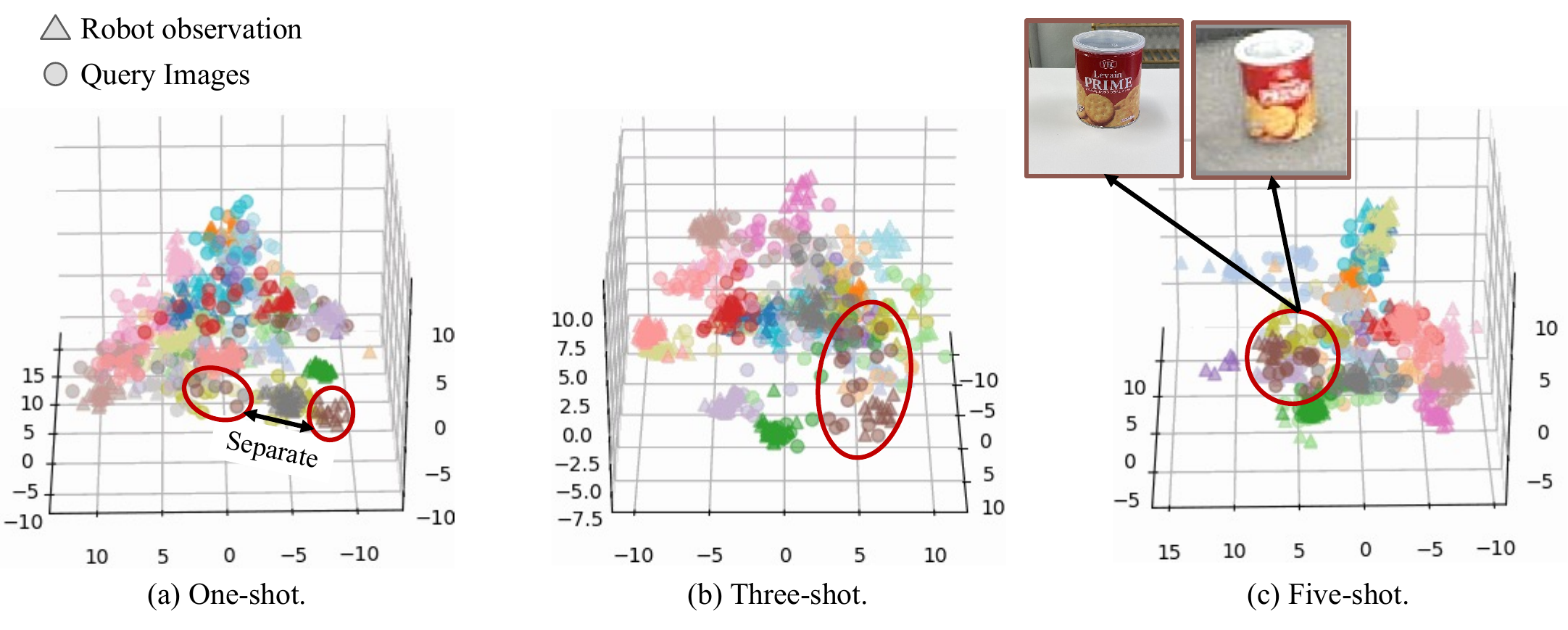}
    \caption{
        % ロボット観察画像と少数撮影の高画質画像を用いて学習させたモデルの、異なる学習条件で得られた潜在空間の比較。
        % 各図の赤い楕円で示すように、学習時に与える高画質画像の枚数が少ないほど、潜在空間において異なる領域間の画像が離れていく。
        Comparison of latent spaces acquired under different training conditions, including those where the model was trained with robot observations and a few high-quality images, reveals the following: As indicated by the red ellipse in each figure, 
        providing more high-quality images during training leads to a closer alignment of images from different domains in the latent space.
        % the fewer high-quality images provided during training, the greater the separation between images from different domains in the latent space.
    }
    \label{fig:ablation_latent_spaces}
\end{figure*}

%%%%%%%%%%%%%%%%%%%%%%%%%%%%%%%%%%%%%%%%%%%%%%%%%%%%%%%%%%%%%%%%%%%%%%%%%%%%%%%%

\subsection{Effectiveness of CrossIA on Baselines}

% 表～ref{tab:all_results}に、表中の条件における評価指標スコアを示す。
% まず、本実験のベースライン法と提案法の評価指標スコアを比較すると、SRスコアが3倍、MRRスコアが2.4倍向上していることがわかる。
% また、ベースライン手法は全タスク試行を通して平均3位でクエリ画像と同じオブジェクトを検索するが、提案手法は平均1位で同じオブジェクトを検索することができる。
% また、SimSiamを用いた条件で、提案手法と評価指標を比較すると、SRスコアは2.3倍、MRRスコアは1.8倍向上する。
% さらに、敵対的学習を加えても、対比学習のみを用いた手法と比較して大きな改善は見られなかった。
Table~\ref{tab:all_results} shows the metric scores under the conditions listed.
Firstly, comparing the evaluation metrics of the baselines SuperGlue and the proposed method, we observe a 2.7-fold improvement in SR and a 2.2-fold improvement in MRR with the proposed method. While SuperGlue typically ranks the matching object third on average across all trials, the proposed method consistently ranks it first.
Additionally, when comparing metrics between the pre-trained SimSiam and the proposed method, the SR score improves by 2.6-fold and the MRR score improves by 2.0-fold with the proposed method. In contrast, SimView showed a much lower success rate due to fine-tuning with only low-quality observation images from the robot.
Furthermore, incorporating adversarial learning into the proposed method did not result in a significant improvement over the version without adversarial learning.

% Table~\ref{tab:all_results} shows the metrics scores under the conditions written in the table.
% First, when we compare the evaluation index scores of the baselines SuperGlue and the proposed method, we discover that the SR score has improved by 3-fold and the MRR score has improved by 2.4-fold.
% SuperGlue searches for the same object as the query image in third place on average across all task trials, but the proposed method can search for the same object in first place on average.
% In addition, when comparing the metrics scores under pre-trained SimSiam and the proposed method, the SR score improves by 2.3-fold and the MRR score improves by 1.8-fold.
% SimView, on the contrary, had a much lower success rate as a result of fine-tuning with only low-quality observation images from the robot.
% Furthermore, the proposed method with the addition of adversarial learning did not show a significant improvement over the method without adversarial learning.
% % Furthermore, adding adversarial learning did not result in significant improvements compared to methods that only use contrastive learning.

%%%%%%%%%%%%%%%%%%%%%%%%%%%%%%%%%%%%%%%%%%%%%%%%%%%%%%%%%%%%%%%%%%%%%%%%%%%%%%%%

\subsection{Effect of Deblurring on Task Success Rate}

% 次に、事前に学習したモデルの条件で、ぼかし除去の有無によるメトリクスのスコアを比較すると、SuperGlue~\cite{superglue}とSimSiam~\cite{simsiam}のいずれの条件でも有意な差は見られない。
% このことから、ロボットの観察画像からぼかしを除去するよりも、領域不変な特徴表現を学習する方がタスク成功率が向上することが示唆される。
When comparing metric scores with and without deblurring in pre-trained models, no significant difference is observed under any conditions.
This suggests that learning domain-invariant feature representations is more effective in improving the task success rate than simply removing blur from the robot's observed images.

%%%%%%%%%%%%%%%%%%%%%%%%%%%%%%%%%%%%%%%%%%%%%%%%%%%%%%%%%%%%%%%%%%%%%%%%%%%%%%%%

\subsection{Qualitative Analysis of Latent Space}

% 最後に、ImageNetで事前学習しただけのSimSiamと、我々の手法で微調整したSimSiamを用いて、潜在空間におけるクエリ画像とロボットが観測した物体画像の分布を比較する。
% Fig.~ref{fig:latent_spaces}(a)に示すように、SimSiamの潜在空間では、高画質画像と低画質画像が潜在空間で分離されていた。
% しかし、提案手法で微調整したSimSiamの潜在空間では、Fig.~ref{fig:latent_spaces}(c)に示すように、ドメインの異なる同一インスタンスのデータが近接して分布している。
% また、ロボットが観測した画像のみでFig.~ref{fig:latent_spaces}(b)を微調整したSimSiamの潜在空間では、Fig.~ref{fig:latent_spaces}(b)に示すように、同一インスタンスの観測画像同士が近接している。
% しかし、高画質画像であるクエリ画像は分離している。
% これらの結果は、提案手法のように、異なるドメインの同一インスタンス画像間で対比学習を行うことで、タスクの成功率が向上することを示唆している。
We compare the distribution of query images ($\circ$) and robot observation images ($\triangle$) in the latent spaces using both pre-trained and fine-tuned SimSiam methods, as shown in Fig.~\ref{fig:latent_spaces}.
In the latent space of the pre-trained SimSiam (Fig.~\ref{fig:latent_spaces}(a)), high-quality and low-quality images are clearly separated.
In the latent space of the Simview, which was trained only with robot observation images, the observed images of the same instance are clustered closely together (Fig.~\ref{fig:latent_spaces}(b)). 
However, the high-quality query images remain distinct from these observed images.
% これは、低品質な画像の識別に過剰適合してしまい、異なるインスタンス同士のクエリ画像が潜在空間上で分離できていないことを示唆している。
This suggests that query images of different instances are not well separated in the latent space, likely due to overfitting to the identification of low-quality images.
In the latent space of CrossIA (Fig.~\ref{fig:latent_spaces}(c)), images of the same instance from different domains are clustered closely together. 
These results indicate that contrastive learning with images of the same instance across different domains, as employed in the proposed method, enhances the task success rate.

% We compare the object distribution of the query images ($\circ$) and the robot observation images ($\triangle$) in the latent spaces using the pre-trained and fine-tuned SimSiam methods, as shown in Fig.~\ref{fig:latent_spaces}.
% In the latent space of the pre-trained SimSiam in Fig.~\ref{fig:latent_spaces}(a), high-quality and low-quality images were separated in the latent space.
% % 
% In addition, in the latent space of the Simview, which was fine-tuned with only images observed by the robot, the observed images of the same instance are close to each other in Fig.~\ref{fig:latent_spaces}(b).
% However, the high-quality query images are separate from the observed images.
% % これは、低品質な画像の識別に過剰適合してしまい、異なるインスタンス同士のクエリ画像が潜在空間上で分離できていないことを示唆している。
% This suggests that the query images between different instances are not separated in the latent space due to over-fitting to the identification of low-quality images.
% % 
% In the latent space of CrossIA in Fig.~\ref{fig:latent_spaces}(c), the data of the same instance with different domains are distributed close to each other.
% These results suggest that contrastive learning between the same instance images of different domains, as in the proposed method, improves the task success rate.

%%%%%%%%%%%%%%%%%%%%%%%%%%%%%%%%%%%%%%%%%%%%%%%%%%%%%%%%%%%%%%%%%%%%%%%%%%%%%%%%

\subsection{Influence of Few-Shot Learning on Performance}

% 表～ref{tab:ablation_all_results}は、学習時に与える高画質画像の枚数を変えた条件でのメトリクススコアである。
% %
% 学習時に与える高画質画像の枚数が少なくなるにつれて、タスクの成功率は低下する。
% Fig.~ref{fig:ablation_latent_spaces}に示すように、学習時に与える高画質画像の枚数が増えるにつれて、異なるドメインの画像は潜在空間において近くなる傾向がある。
% この結果は、訓練時に十分な高画質画像を与えることで、対比学習によるドメイン不変な特徴表現の獲得能力が向上することを示唆している。
Table~\ref{tab:ablation_all_results} shows the metrics scores for a varying number of high-quality images during learning.
The task success rate decreases as the number of high-quality images provided during training decreases.
As shown in Fig.~\ref{fig:ablation_latent_spaces}, images from different domains tend to cluster more closely in the latent space as the number of high-quality images increases. 
This result suggests that providing a sufficient number of high-quality images during training improves the ability to learn domain-invariant feature representations through contrastive learning. 

% % To verify the effect of providing a few high-quality images during fine-tuning, we examined the improvement in task success rates and how it decreases as fewer images are provided for each object.
% Table~\ref{tab:ablation_all_results} shows the metrics scores under conditions where the number of high-quality images given during learning was changed.
% %
% The success rate of the task decreases as the number of high-quality images given during training decreases.
% As shown in Fig.~\ref{fig:ablation_latent_spaces}, images from different domains tend to become closer in the latent space as the number of high-quality images provided during training increases. 
% This result suggests that providing adequate high-quality images during training enhances the ability to obtain domain-invariant feature representations through contrastive learning. 
% % Therefore, ensuring adequate high-quality images is essential for achieving domain-invariant feature representation.

% ワンショット条件下において、敵対的学習を加えると、対照的学習のみと比較して、全ての評価指標のスコアが向上した。
% この結果は、敵対的学習と対比的学習を組み合わせたマルチタスク学習は、対比的学習のみと比較して、高品質な画像が少なくても、異なる領域の画像を潜在空間に近づけることができることを示唆している。
Under one-shot conditions, adding adversarial learning improved all evaluation metric scores compared to using only contrastive learning, as shown in Table~\ref{tab:ablation_all_results}.
This result suggests that combining adversarial and contrastive learning can bring images from different domains closer together in the latent space, even with fewer high-quality images than when using contrastive learning alone.

% Under one-shot conditions, when adversarial learning was added, the scores of all evaluation metrics improved compared to only contrastive learning in Table~\ref{tab:ablation_all_results}.
% This result suggests that multi-task learning of adversarial and contrastive learning could make images of different domains closer to the latent space even with fewer high-quality images than only contrastive learning.

%%%%%%%%%%%%%%%%%%%%%%%%%%%%%%%%%%%%%%%%%%%%%%%%%%%%%%%%%%%%%%%%%%%%%%%%%%%%%%%%
% Conclusion
%%%%%%%%%%%%%%%%%%%%%%%%%%%%%%%%%%%%%%%%%%%%%%%%%%%%%%%%%%%%%%%%%%%%%%%%%%%%%%%%

\section{Conclusion}
\label{sec:conclusion}

% 本研究では、InstanceImageNavにおいて、ロボットが観測した画像とユーザが提供した画像との間の領域ギャップに起因する成功率の低下の問題に対処する手法を提案した。本手法では、数枚の高画質画像とロボット観測画像を利用することで、このギャップを緩和する。
% 具体的には、ロボットが観測した画像と高画質数ショット画像の間で、領域不変な特徴表現を学習する対比学習を用いた手法CrossIAを提案した。
% 提案手法とベースライン手法を実験により比較した。
% 実験の結果、ベースライン法と比較して、タスクの成功率が3倍向上することが示された。
In this study, we proposed a method to address the reduced success rates in InstanceImageNav due to the domain gap between robot-observed images and user-provided images. 
Our approach leverages a few high-quality images along with robot observations to bridge this gap. 
Specifically, we introduced CrossIA, a method that employs contrastive learning to develop domain-invariant feature representations between robot-observed images and high-quality few-shot images.
Experimental results demonstrated that the task success rate improved three-fold compared to SuperGlue. 
Additionally, the impact of deblurring on task success rates was found to be limited.

% limitation
% 提案手法の限界は、ユーザが各オブジェクトに対して少数の高画質画像を提供することに依存していることである。本手法は少数の画像で効果的に機能するように設計されているが、検索する対象物の数が増えるにつれて、各対象物に対する高画質画像の要求も大きくなる。これは、多数の物体についてそのような画像を得ることが困難な実世界のシナリオでは非現実的となる可能性がある。したがって、今後の研究では、自動画像強調や合成データ生成などの技術を活用することで、ユーザーが提供する高画質画像への依存を排除または低減できる手法の開発に焦点を当てるべきである。
A limitation of the proposed method is its reliance on users providing a few high-quality images for each object. 
Although the method is designed to work effectively with a small number of images, the need for high-quality images increases as the number of objects to be searched grows. 
This can become impractical in real-world scenarios where obtaining such images for many objects is challenging.
Future work should focus on developing methods to reduce or eliminate the dependency on user-provided high-quality images.

Future work will explore the use of advanced image restoration methods, such as pre-trained diffusion models~\cite{diffbir}, to automatically generate high-quality images from low-quality images captured by robots. 
This approach could improve the success rate of InstanceImageNav, especially in real-world scenarios with limited high-quality images.

%%%%%%%%%%%%%%%%%%%%%%%%%%%%%%%%%%%%%%%%%%%%%%%%%%%%%%%%%%%%%%%%%%%%%%%%%%%%%%%%
% Appendix
%%%%%%%%%%%%%%%%%%%%%%%%%%%%%%%%%%%%%%%%%%%%%%%%%%%%%%%%%%%%%%%%%%%%%%%%%%%%%%%%

%\section*{Appendix}

%%%%%%%%%%%%%%%%%%%%%%%%%%%%%%%%%%%%%%%%%%%%%%%%%%%%%%%%%%%%%%%%%%%%%%%%%%%%%%%%
% Acknowledgment
%%%%%%%%%%%%%%%%%%%%%%%%%%%%%%%%%%%%%%%%%%%%%%%%%%%%%%%%%%%%%%%%%%%%%%%%%%%%%%%%

%\section*{Acknowledgment}

%%%%%%%%%%%%%%%%%%%%%%%%%%%%%%%%%%%%%%%%%%%%%%%%%%%%%%%%%%%%%%%%%%%%%%%%%%%%%%%%
% Bibliography
%%%%%%%%%%%%%%%%%%%%%%%%%%%%%%%%%%%%%%%%%%%%%%%%%%%%%%%%%%%%%%%%%%%%%%%%%%%%%%%%

\bibliographystyle{templates/IEEEtran}
\bibliography{root}

%%%%%%%%%%%%%%%%%%%%%%%%%%%%%%%%%%%%%%%%%%%%%%%%%%%%%%%%%%%%%%%%%%%%%%%%%%%%%%%%

%\balance

%%%%%%%%%%%%%%%%%%%%%%%%%%%%%%%%%%%%%%%%%%%%%%%%%%%%%%%%%%%%%%%%%%%%%%%%%%%%%%%%

\end{document}